\documentclass{article} % For LaTeX2e
\usepackage{iclr2024_conference,times}

\newtheorem{theorem}{Theorem}[section]

\usepackage{xspace}
\usepackage{xcolor}
\usepackage{overpic}
\usepackage{amsmath,amssymb,amsfonts,dsfont,pifont,bm,bbm,mathrsfs,mathtools,nicefrac}
\usepackage{algorithm,algpseudocode,listings}
\usepackage{booktabs,multirow,adjustbox,diagbox,threeparttable}
\usepackage{graphicx, caption}
\usepackage{wrapfig}

\usepackage{amsmath,amsfonts,bm}

\def\eqref#1{equation~\ref{#1}}
\def\1{\bm{1}}

\def\dd{{\rm d}}
\def\defeq{{:=}}
\def\bvx{{\bm{\rvx}}}
\def\beps{{\bm{\epsilon}}}

\def\rvepsilon{{\mathbf{\epsilon}}}

\def\rvw{{\mathbf{w}}}
\def\rvx{{\mathbf{x}}}

\def\rvz{{\mathbf{z}}}

\DeclareMathAlphabet{\mathsfit}{\encodingdefault}{\sfdefault}{m}{sl}
\SetMathAlphabet{\mathsfit}{bold}{\encodingdefault}{\sfdefault}{bx}{n}

\def\gN{{\mathcal{N}}}

\def\gU{{\mathcal{U}}}

\def\sT{{\mathbb{T}}}

\DeclareMathOperator*{\argmin}{arg\,min}

\newcommand{\tocite}[1]{\textcolor{red}{[TO CITE]}}

\newcommand{\methodabbr}{E-TSDM\xspace}  % Timestep Shared Diffusion Model

\newcommand\nonumfootnote[1]{%
\begingroup%
    \renewcommand\thefootnote{}\footnote{\hspace{-4pt}#1}%
    \addtocounter{footnote}{-1}%
\endgroup%
}

\definecolor{citeblue}{RGB}{48,111,186}
\definecolor{modifiedblue}{RGB}{28,91,206}
\definecolor{deletered}{RGB}{206,91,28}
\usepackage[pagebackref=false,breaklinks=true,colorlinks=true,citecolor=citeblue,bookmarks=false]{hyperref}
\usepackage{graphicx}
\usepackage{subfigure}
\usepackage[capitalize,noabbrev]{cleveref}

\title{Lipschitz Singularities in Diffusion Models}

\author{
    Zhantao Yang$^{1,4}$$\star$ \quad
    Ruili Feng$^{2,4}$$\diamond$ \quad
    Han Zhang$^{1,4}$$\star$ \quad
    Yujun Shen$^{3}$ \quad
    Kai Zhu$^{2,4}$ \And
    Lianghua Huang$^{4}$ \quad
    Yifei Zhang$^{1,4}$$\star$ \quad
    Yu Liu$^{4}$ \quad
    Deli Zhao$^{4}$ \quad
    Jingren Zhou$^{4}$ \quad
    Fan Cheng$^{1}$$\dagger$\\
    \\[5pt]
    $^1$Shanghai Jiao Tong University \quad \\
    $^2$University of Science and Technology of China \quad 
    $^3$Ant Group \quad
    $^4$Alibaba Group 
    \\[5pt]
    \footnotesize\texttt{\{ztyang196, ruilifengustc, hzhang9617, shenyujun0302\}@gmail.com} \quad
    \\[0pt]
    \footnotesize\texttt{\{zkzy\}@mail.ustc.edu.cn} \quad
    \footnotesize\texttt{\{xuangen.hlh\}@alibaba-inc.com}
    \\[0pt]
    \footnotesize\texttt{qidouxiong619@sjtu.edu.cn} \quad
    \footnotesize\texttt{\{ly103369\}@alibaba-inc.com}
    \\[0pt]
    \footnotesize\texttt{zhaodeli@gmail.com} \quad
    \footnotesize\texttt{jingren.zhou@alibaba-inc.com} \quad
    \footnotesize\texttt{chengfan@sjtu.edu.cn}
}

\iclrfinalcopy % Uncomment for camera-ready version, but NOT for submission.
\begin{document}
\maketitle

\begin{abstract}
Diffusion models, which employ stochastic differential equations to sample images through integrals, have emerged as a dominant class of generative models.
However, the rationality of the diffusion process itself receives limited attention, leaving the question of whether the problem is well-posed and well-conditioned.
In this paper, we explore a perplexing tendency of diffusion models: they often display the infinite Lipschitz property of the network with respect to time variable near the zero point.
We provide theoretical proofs to illustrate the presence of infinite Lipschitz constants and empirical results to confirm it.
The Lipschitz singularities pose a threat to the stability and accuracy during both the training and inference processes of diffusion models.
Therefore, the mitigation of Lipschitz singularities holds great potential
for enhancing the performance of diffusion models.
To address this challenge, we propose a novel approach, dubbed \methodabbr, which alleviates the Lipschitz singularities of the diffusion model near the zero point of timesteps.
Remarkably, our technique yields a substantial improvement in performance.
Moreover, as a byproduct of our method, we achieve a dramatic reduction in the Fréchet Inception Distance of acceleration methods relying on network Lipschitz, including DDIM and DPM-Solver, by over 33\%.
Extensive experiments on diverse datasets validate our theory and method.
Our work may advance the understanding of the general diffusion process, and also provide insights for the design of diffusion models.

\nonumfootnote{$\dagger$ Corresponding author, $\star$ Work performed at Alibaba Academy, $\diamond$ Project leader}

\end{abstract}

\vspace{-20pt}
\section{Introduction}\label{sec:intro}

The rapid development of diffusion models has been witnessed in image synthesis~\citep{ho2020denoising, song2020denoising, ramesh2022hierarchical, saharia2022photorealistic,rombach2022high,zhang2023adding,hoogeboom2023simple} in the past few years.
Concretely, diffusion models construct a multi-step process to destroy a signal by gradually adding noises to it.
That way, reversing the diffusion process (\textit{i.e.}, denoising) at each step naturally admits a sampling capability.
In essence, the sampling process involves solving a reverse-time stochastic differential equation (SDE) through integrals~\citep{song2020score}.

Although diffusion models have achieved great success in image synthesis, the rationality of the diffusion process itself has received limited attention, leaving the open question of whether the problem is well-posed and well-conditioned.
In this paper, we surprisingly observe that the noise-prediction~\citep{ho2020denoising} and v-prediction~\citep{salimans2022progressive} diffusion models often exhibit a perplexing tendency to possess infinite Lipschitz of network with respect to time variable near the zero point.
We provide theoretical proofs to illustrate the presence of infinite Lipschitz constants and empirical results to confirm it.
Given that noise prediction and v-prediction are widely adopted by popular diffusion models~\citep{dhariwal2021diffusion, rombach2022high, ramesh2022hierarchical, saharia2022photorealistic, podell2023sdxl}, the presence of large Lipschitz constants is a significant problem for the diffusion model community.

Since we uniformly sample timesteps for both training and inference processes, large Lipschitz constants \textit{w.r.t.} time variable pose a significant threat to both training and inference processes of diffusion models.
When training, large Lipschitz constants near the zero point affect the training of other parts due to the smooth nature of the network, resulting in instability and inaccuracy.
Moreover, since inference requires a smooth network for integration, the large Lipschitz constants probably have a substantial impact on accuracy, particularly for faster samplers.
Therefore, the mitigation of Lipschitz singularities holds great potential for enhancing the performance of diffusion models.

\begin{figure*}[t]
\begin{center}
\begin{overpic}[width=1.0\linewidth]{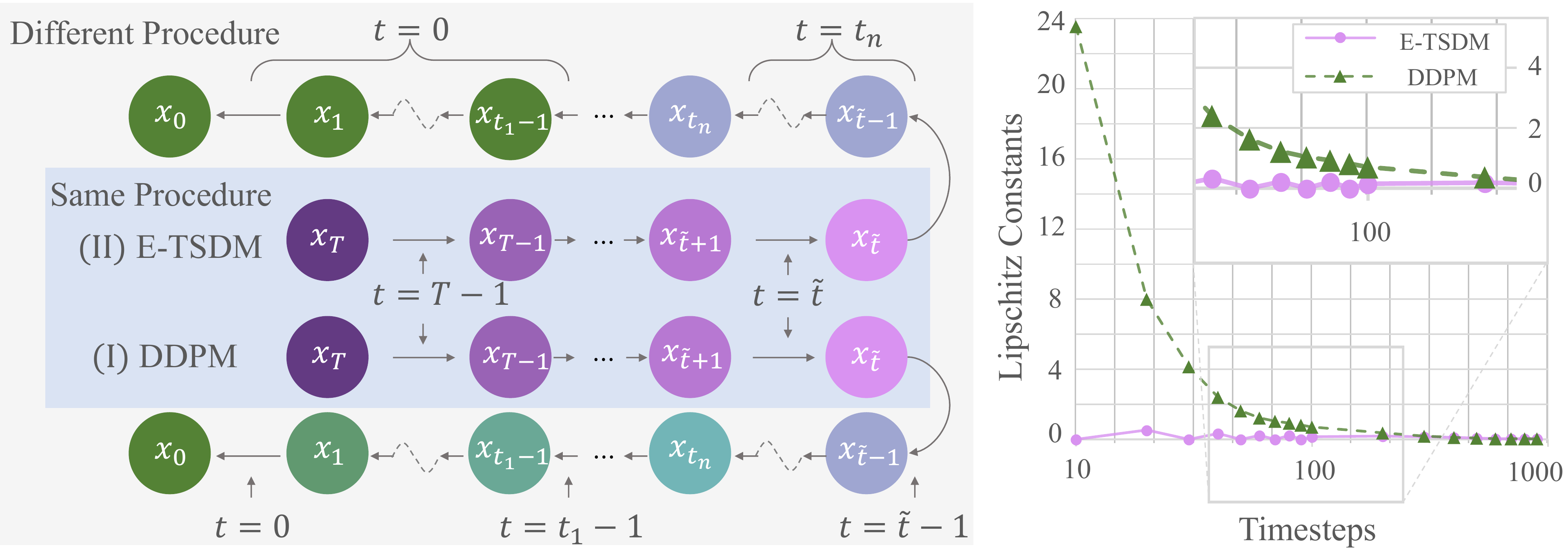}
% \put(75,5.5){\textbf{$\tilde{t}$ (=100) }}
% \put(78,0){Timesteps}
\put(18,-4){(a) Conceptual comparison}
\put(70,-4){(b) Lipschitz constants}
% \put(63, 8){\rotatebox{90}{Lipschitz Constants}}
\end{overpic}
\end{center}
\vspace{10pt}
\caption{
    (a) \textbf{Conceptual comparison} between DDPM~\citep{ho2020denoising} (\uppercase\expandafter{\romannumeral1}) and our proposed early timestep-shared diffusion model (\methodabbr) (\uppercase\expandafter{\romannumeral2}).
    DDPM trains the network $\epsilon_\theta(\cdot, t)$ with varying timestep conditions $t$ at each denoising step, whereas \methodabbr uniformly divides the near-zero timestep interval $t\in [0, \tilde{t})$ with high Lipschitz constants into $n$ sub-intervals and \textbf{shares the condition $t$ within each sub-interval}.
    Here, $\tilde{t}$ denotes the length of the interval for sharing conditions.
    When $t \ge \tilde{t}$, \methodabbr follows the same procedure as DDPM.
    However, when $t < \tilde{t}$, \methodabbr shares timestep conditions.
    (b) \textbf{Quantitative comparison} of the Lipschitz constants between DDPM and our proposed early timestep-shared diffusion model (\methodabbr).
    \textbf{The Lipschitz constants tend to be extremely large near zero point for DDPM}.
    However, our sharing approach allows \methodabbr to force the Lipschitz constants in each sub-interval to be zero, thereby \textbf{reducing the overall Lipschitz constants} in the timestep interval $t\in [0, \tilde{t})$, where $\tilde{t}$ is set as a default value 100.
}
\label{fig:framework}
\vspace{-19pt}
\end{figure*}

Fortunately, there is a simple yet effective alternative solution: by sharing the timestep conditions in the interval with large Lipschitz constants, the Lipschitz constants can be set to zero.
Based on this idea, we propose a practical approach, which uniformly divides the target interval near the zero point into $n$ sub-intervals, and uses the same condition values in each sub-interval, as shown in \cref{fig:framework} (\uppercase\expandafter{\romannumeral2}).
By doing so, this approach can effectively reduce the Lipschitz constants near $t=0$ to zero.
To validate this idea, we conduct extensive experiments, including unconditional generation on various datasets, acceleration of sampling, and super-resolution task.
Both qualitative and quantitative results confirm that our approach substantially alleviates the large Lipschitz constants near zero point and improves the synthesis performance compared to the DDPM baseline~\citep{ho2020denoising}.
We also compare this simple approach with other potential methods to address the challenge of large Lipschitz constants, and find our method outperforms all of these alternative methods.
In conclusion, in this work, we theoretically prove and empirically observe the presence of Lipschitz singularities issue near the zero point, advancing the understanding of the diffusion process.
Besides, we propose a simple yet effective approach to address this challenge and achieve impressive improvements.

\section{Related work}\label{sec:related}

The significant advancements of diffusion models have been witnessed in recent years in the domain of image generation.~\citep{karras2022elucidating, lu2022dpm, dockhorn2021score, bao2022analytic, lu2022maximum, bao2022estimating, zhang2023dimensionality}.
It~\citep{sohl2015deep, ho2020denoising, song2020score} defines a Markovian forward process $\{\bvx_t\}_{t\in [0, T]}$ that gradually destroys the data $\rvx_0$ with Gaussian noise.
For any $t\in [0, T]$, the conditional distribution $q_{0t}(\rvx_t|\rvx_0)$ satisfies
\begin{equation}
\label{eq:diffusion-one-step-forward}
    q_{0t}\left(\rvx_t | \rvx_0\right) = \gN\left(\rvx_t | \alpha_t \rvx_0, \sigma_t^2 \mathbf{I}\right),
\end{equation}
where $\alpha_t$ and $\sigma_t$ are referred to as the noise schedule, satisfying $\alpha_t^2 + \sigma_t^2 = 1$.
Generally, $\alpha_t$ decreases from $1$ to $0$ as $t$ increases, to ensure that the marginal distribution of $\rvx_t$ gradually changes from the data distribution $q_0(x_0)$ to Gaussian. 
\citet{kingma2021variational} further prove that the following stochastic differential equation (SDE) has the same transition distribution $q_{0t}(\rvx_t|\rvx_0)$ as in \cref{eq:diffusion-one-step-forward}
for any $t\in [0, T]$:
\begin{equation}
\label{eq:diffusion-forward-SDE}
    \dd \rvx_t = f\left(t\right) \rvx_t \dd t + g\left(t\right) \dd \rvw_t,\quad \rvx_0 \sim q_0\left(x_0\right),
\end{equation}
where $\rvw_t$ is the standard Wiener process, $f(t) = \frac{\dd \log \alpha_t}{\dd t}$ and $g(t) = 2\sigma_t^2 \frac{\dd \log (\sigma_t / \alpha_t)}{\dd t}$.

\citet{song2020score} point out that the following reverse-time SDE has the same marginal distribution $q_t(\rvx_t)$ for any $t\in [0, T]$: 
\begin{equation}
\label{eq:diffusion-reverse-SDE}
    \dd \rvx_t = [f\left(t\right)\rvx_t - g\left(t\right)^2 \nabla_{\rvx_t}\log q_t\left(\rvx_t\right)] \dd t + g\left(t\right) \dd \bar{\rvw}_t,\quad \rvx_T \sim q_T\left(\rvx_T\right),
\end{equation}
where $\bar{\rvw}_t$ is a standard Wiener process in the reverse time. Once the score function $\nabla_{\rvx_t}\log q_t(\rvx_t)$ is known, we can simulate \cref{eq:diffusion-reverse-SDE} for sampling.
However, directly learning the score function is problematic, as it involves an explosion of training loss when having a small $\sigma_t$\citep{song2020score}.
In practice, the noise prediction model $\beps_\theta(\rvx_t, t)$ is often adopted to estimate $-\sigma_t \nabla_{\rvx_t}\log q_t(\rvx_t)$.
The network $\beps_\theta(\rvx_t,t)$ can be trained by minimizing the objective:
\begin{equation}
\label{eq:ddpm-loss}
    \displaystyle
    \mathcal{L}\left(\theta\right):= \mathbb{E}_{t \sim \mathcal{U}\left(0,T\right), \rvx_0\sim q_0\left(\rvx_0\right), \rvepsilon \sim \gN\left(0, \mathbf{I}\right)} \left[  \Vert \beps_\theta\left(\alpha_t \rvx_0 + \sigma_t \rvepsilon, t\right)-\rvepsilon \Vert _2^2\right].
\end{equation}
In this work, our observation of Lipschitz singularities on noise-prediction and v-prediction diffusion models reveals the inherent price of such an approach.

\noindent\textbf{Numerical stability near zero point.}
% [CHATGPT]
Achieving numerical stability is essential for high-quality samples in diffusion models, where the sampling process involves solving a reverse-time SDE.
\begin{wrapfigure}{r}{0.48\textwidth}
\vspace{-12pt}
\begin{center}
% \centering
% \begin{overpic}[width=0.45\linewidth]{figures_v2/stability.pdf}
\includegraphics[width=0.48\textwidth]{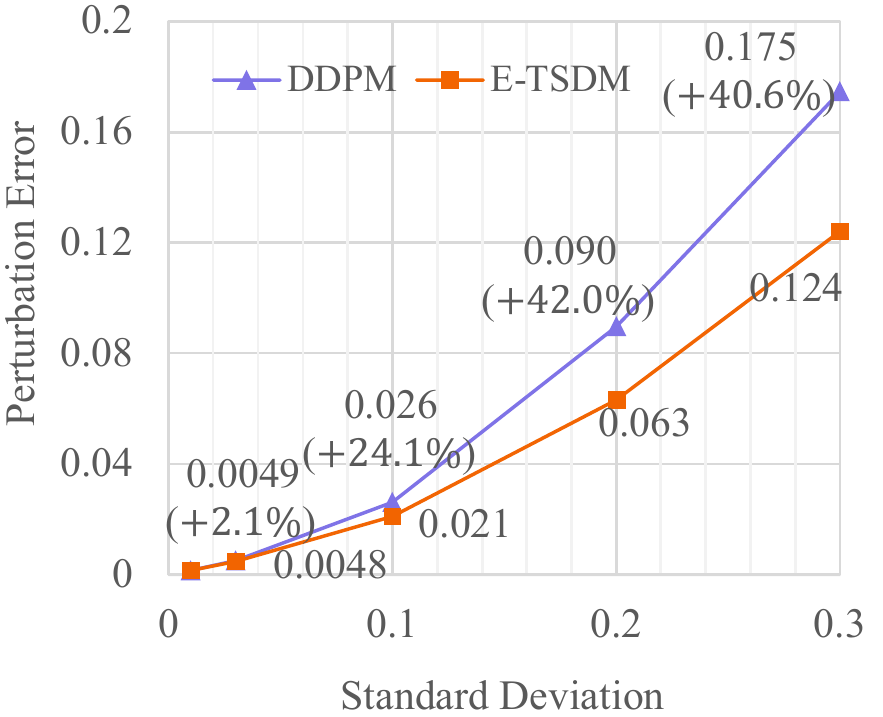}
\end{center}
\vspace{-12pt}
\caption{\textbf{Quantitative comparison} of the errors caused by a perturbation on the input between \methodabbr and DDPM~\citep{ho2020denoising}.
Results show that \textbf{\methodabbr is more stable}, as its prediction is less affected, 
\textit{e.g.}, the perturbation error of DDPM is 42.0\% larger than \methodabbr when the perturbation scale is 0.2.}
\label{fig:stability}
\vspace{-12pt}
\end{wrapfigure}
Nevertheless, numerical instability is frequently observed near $t=0$ in practice~\citep{song2021maximum,vahdat2021score}.
To address this singularity, one possible approach is to set a small non-zero starting time $\tau>0$ in both training and inference~\citep{song2021maximum,vahdat2021score}.
~\citet{kim2022soft} resolve the trade-off between density estimation and sample generation performance by introducing randomization to the fixed $\tau$.
In contrast, we enhance numerical stability by reducing the Lipschitz constants to zero near $t=0$, which leads to improved sample quality in diffusion models.
It is worth noting that the numerical issues observed by aforementioned works are mainly caused by the singularity of transition kernel $q_{0t}(\rvx_t|\rvx_0)$.
This transition kernel will degrade to a Dirac kernel $\delta(\rvx_t - \alpha_t \rvx_0)$ as $\sigma_t \rightarrow 0$.
However, our observation is \textit{the infinite Lipschitz constants of the noise prediction model $\beps_\theta\left(\rvx, t\right)$ w.r.t time variable $t$}, and this is caused by the explosion of $\frac{\dd \sigma_t}{\dd t}$ as $t \rightarrow 0$.
To the best of our knowledge, this has not been observed before.

\section{Lipschitz singularities in diffusion models}\label{sec:analyze}

\noindent\textbf{Lipschitz singularities issue.}
In this section, we elucidate the vexing propensity of diffusion models to exhibit infinite Lipschitz near the zero point.
We achieve this by analyzing the partial derivative $\partial \beps_\theta(\rvx, t)/\partial t$ of the network $\beps_\theta(\rvx, t)$.
In essence, the emergence of Lipschitz singularities, characterized by $\lim \sup_{t\rightarrow 0+} \left\lVert \frac{\partial \beps_\theta(\rvx, t)}{\partial t} \right\rVert \rightarrow \infty$, can be attributed to the fact that the prevailing noise schedules conform to the behavior of $d \sigma_t / dt \rightarrow \infty$ as the parameter $t$ tends towards zero.

\noindent\textbf{Theoretical analysis.}
Now we theoretically prove that the infinite Lipschitz happens near the zero point in diffusion models, where the distribution of data is an arbitrary complex distribution.
We focus particularly on the scenario where the network $\beps_\theta(\rvx, t)$ is trained to predict the noises added to images (v-prediction model~\citep{salimans2022progressive} has a similar singularity problem, and is analyzed in \cref{subsec:v-prediction}).
The network $\beps_\theta(\rvx, t)$ exhibits a relationship with the score function $\nabla_\rvx \log q_t(\rvx)$ that $\beps_\theta(\rvx, t) = - \sigma_t \nabla_\rvx \log q_t(\rvx)$~\citep{song2020score},
where $\sigma_t$ is the standard deviation of the forward transition distribution $q_{0t}(\rvx|\rvx_0)=\mathcal{N}(\rvx; \alpha_t \rvx_0, \sigma_t^2 \mathbf{I})$.
Specifically, $\alpha_t$ and $\sigma_t$ satisfy $\alpha_t^2 + \sigma_t^2 = 1$.
\begin{theorem}
\vspace{-3pt}
\label{prop:xt-diff}
Given a noise schedule, since $\sigma_t = \sqrt{1 - \alpha_t^2}$, we have $\frac{d \sigma_t}{dt} = -\frac{\alpha_t}{\sqrt{1-\alpha_t^2}} \frac{d\alpha_t}{dt}$.
As $t$ gets close to 0, the noise schedule requires $\alpha_t \rightarrow 1$, leading to $d \sigma_t / dt \rightarrow \infty$ as long as $\frac{d\alpha_t}{dt}|_{t=0}\neq 0$.
The partial derivative of the network can be written as 
\begin{equation}
% \begin{aligned}
\label{eq:partial-derivative}
    \dfrac{\partial \beps_\theta\left(\rvx, t\right)}{\partial t} = \dfrac{\alpha_t}{\sqrt{1-\alpha_t^2}} \dfrac{d\alpha_t}{dt} \nabla_\rvx \log q_t\left(\rvx\right) - \dfrac{\partial \nabla_\rvx \log q_t\left(\rvx\right)}{\partial t}\sigma_t.
% \end{aligned}
\end{equation}
Note that $\alpha_t \rightarrow 1$ as $t \rightarrow 0$, thus if $\frac{d\alpha_t}{dt}|_{t=0}\neq 0$, and $\nabla_\rvx \log q_t(\rvx)|_{t=0}\neq \mathbf{0}$, then one of the following two must stand
\begin{equation}
\label{eq:partial-derivative-condition}
    \lim \sup_{t\rightarrow 0+} \left\lVert \dfrac{\partial \beps_\theta\left(\rvx, t\right)}{\partial t} \right\rVert \rightarrow \infty ; \quad \lim \sup_{t\rightarrow 0+} \left\lVert \dfrac{\partial \nabla_\rvx \log q_t\left(\rvx\right)}{\partial t}\sigma_t \right\rVert \rightarrow \infty.
\end{equation}
\end{theorem}
Note that $\frac{d\alpha_t}{dt}|_{t=0}\neq 0$ stands for a wide range of noise schedules, including linear, cosine, and quadratic schedules (see details in \cref{subsec:alpha-derivatives}).
Besides, we can safely assume that $q_t(\rvx)$ is a smooth process.
Therefore, we may often have $\lim \sup_{t\rightarrow 0+} \big\lVert \frac{\partial \beps_\theta(\rvx, t)}{\partial t} \big\rVert \rightarrow \infty$, indicating the infinite Lipschitz constants around $t=0$.

\noindent\textbf{Simple case illustration.}
Take a simple case that the distribution of data $p(\rvx_0) \sim \mathcal{N}(\mathbf{0}, \mathbf{I})$ for instance, the score function for any $t\in [0, T]$ can be written as
\begin{equation}
\label{eq:lipschitz-continuous-gaussian}
    \nabla_\rvx \log q_t\left(\rvx\right) = \nabla_\rvx \log \left(\dfrac{1}{\sqrt{2\pi}}\exp \left(-\dfrac{\Vert \rvx \Vert_2^2}{2}\right)\right) = - \rvx.
\end{equation}
Due to the relationship $\beps_\theta(\rvx, t) = - \sigma_t \nabla_\rvx \log q_t(\rvx)$ and the fact that the deviation $\frac{{\rm d} \sigma_t}{{\rm d} t}$ tends toward $\infty$ as $t\rightarrow 0$, we have $\big\lVert \frac{\partial \beps_\theta(\rvx, t)}{\partial t} \big\rVert \rightarrow \infty$.

\noindent\textbf{Case in reality.}
After theoretically proving that diffusion models suffer infinite Lipschitz near the zero
point, we show it empirically. 
We estimate the Lipschitz constants of a network by
\begin{equation}
\label{eq:lipschitz-constant}
    K(t, t^\prime) = \dfrac{\mathbb{E}_{\rvx_t}[\lVert \beps_\theta\left(\rvx_t, t\right)-\beps_\theta\left(\rvx_t, t^\prime\right)] \lVert_2]}{\Delta t},
\end{equation}
where $\Delta t = \lvert t - t^\prime \rvert$.
For a network $\beps_\theta(\rvx_t, t^\prime)$ of DDPM baseline~\citep{ho2020denoising} trained on FFHQ $256\times256$~\citep{karras2019style} (see training details in \cref{subsec:settings} and more results of the Lipschitz constants $K(t, t^\prime)$ on other datasets in \cref{sec:app-results:lipschitz-constants}), the variation of the Lipschitz constants $K(t, t^\prime)$ as the noise level $t$ varies is seen in \cref{fig:framework}b, showing that the Lipschitz constants $K(t, t^\prime)$ get extremely large in the interval with low noise levels.
Such large Lipschitz constants support the above theoretical analysis and pose a threat to the stability and accuracy of the diffusion process, which relies on integral operations.
\section{Mitigating Lipschitz singularities by sharing conditions}\label{sec:method}

\noindent\textbf{Proposed method.}
In this section, we propose the Early Timestep-shared Diffusion Model (\methodabbr), which aims to alleviate the Lipschitz singularities by sharing the timestep conditions in the interval with large Lipschitz constants.
To avoid impairing the network's ability, \methodabbr performs a stepwise operation of sharing timestep condition values. 
Specifically, we consider the interval near the zero point suffering from large Lipschitz constants, denoted as $[0, \tilde{t})$, where $\tilde{t}$ indicates the length of the target interval.
\methodabbr uniformly divides this interval into $n$ sub-intervals represented as a sequence $\sT=\{t_0, t_1, \cdots, t_n\}$, where $0=t_0<t_1<\dots<t_n=\tilde{t}$ and $t_1 - t_0 = t_{i} - t_{i-1}, \forall i = 1,2,\cdots,n$.
For each sub-interval, \methodabbr employs a single timestep value (the left endpoint of the sub-interval) as the condition, both during training and inference.
Utilizing this strategy, \methodabbr effectively enforces zero Lipschitz constants within each sub-interval, with only the timesteps located near the boundaries of the sub-intervals having a Lipschitz constant greater than zero.
As a result, the overall Lipschitz constants of the target interval $t\in [0, \tilde{t})$ are significantly reduced.
The corresponding training loss can be written as
\begin{equation}
\label{eq:loss}
    \displaystyle
    \mathcal{L}\left(\rvepsilon_\theta\right):= \mathbb{E}_{t \sim \mathcal{U}\left(0,T\right), \rvx_0\sim q\left(\rvx_0\right), \rvepsilon \sim \gN\left(0, \mathbf{I}\right)}\left[ \Vert \rvepsilon_\theta\left(\alpha_t\rvx_0+\sigma_t\rvepsilon, f_{\sT}\left(t\right)\right)-\rvepsilon \Vert _2^2\right],
\end{equation}

where $f_{\sT}(t)=\max_{1\le i \le n}\{t_{i-1}\in\sT: t_{i-1} \le t\}$ for $t < \tilde{t}$, while $f_{\sT}(t)=t$ for $t \ge \tilde{t}$.
The corresponding reverse process can be represented as
\begin{equation}
\label{eq:transition-kernel}
    \displaystyle
    p_\theta\left(\rvx_{t-1}|\rvx_t\right) = \mathcal{N}\left(\rvx_{t-1};\frac{\alpha_{t-1}}{\alpha_t} \left(\rvx_t -\frac{\beta_t}{\sigma_t}\rvepsilon_\theta\left(\rvx_t, f_{\sT}\left(t\right)\right)\right), \eta_t^2 \mathbf{I}\right),
\end{equation}
where $\beta_t = 1 - \frac{\alpha_t}{\alpha_{t-1}}$, and $\eta_t^2 = \beta_t$.
\methodabbr is easy to implement, and the algorithm details are provided in \cref{subsec:app-code}.

\begin{figure*}[t]
\begin{center}
\begin{overpic}[width=1.0\linewidth]{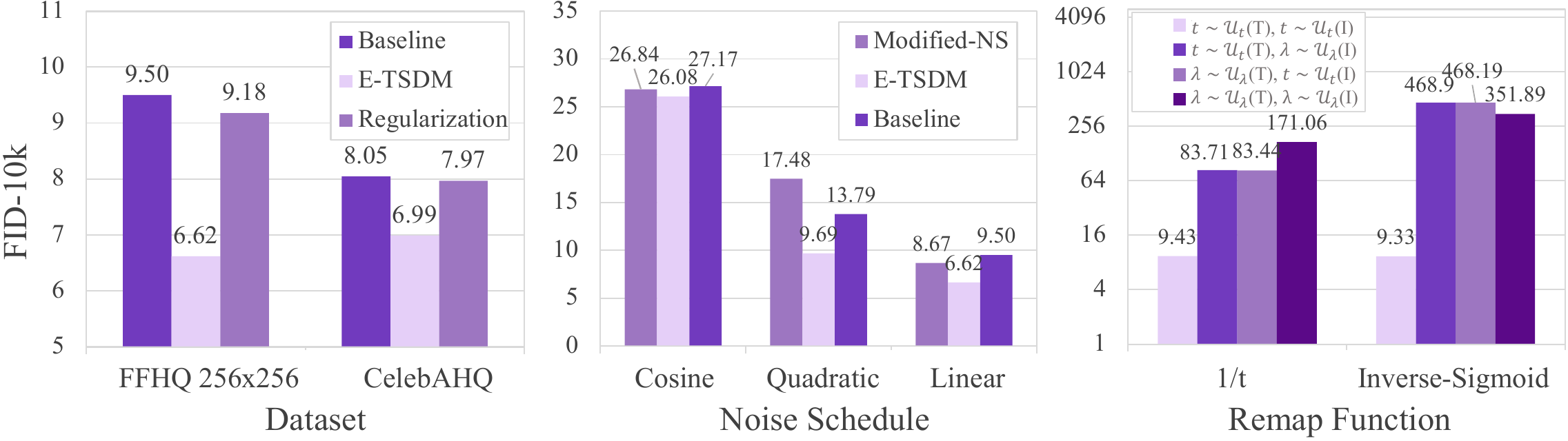}
% \put(75,5.5){\textbf{$\tilde{t}$ (=100) }}
% \put(78,0){Timesteps}
\put(10,-2){(a) Regularization}
\put(35,-2){(b) Modification of noise schedules}
\put(80,-2){(c) Remap}
% \put(63, 8){\rotatebox{90}{Lipschitz Constants}}
\end{overpic}
\end{center}
\vspace{2pt}
\caption{
    \textbf{Quantitative analysis of alternative methods} evaluated with FID-10k $\downarrow$. (a) \textbf{Regularization}: Experimental results on FFHQ $256\times256$ and CelebAHQ $256\times256$ show that \textbf{regularization techniques can slightly improve the FID of DDPM~\citep{ho2020denoising} baseline but performs worse than \methodabbr}
    (b) \textbf{Modification of noise schedules (Modified-NS)}: We implement Modified-NS on linear, quadratic, and cosine schedules. 
    Experimental results on FFHQ $256\times256$ dataset indicate that the performance of \textbf{Modified-NS is unstable while \methodabbr achieves better synthesis performance}.
    (c) \textbf{Remap}: \textbf{Quantitative comparison} of \textbf{remap} method between uniformly sampling $t$ and uniformly sampling $\lambda$, during training and inference, on FFHQ $256\times256$.
    Specifically, $\mathcal{U}_t$ is $\mathcal{U}[0,1]$, and $\mathcal{U}_\lambda$ is $\mathcal{U}[0, K]$ for $1/t$ but $\mathcal{U}[-K, K]$ for Inverse-Sigmoid, where $K$ is a large number to avoid infinity.
    (T) represents the sampling strategy during the training process while (I) represents that during the inference process.
    Results show that \textbf{remap is not helpful}.
}
\label{fig:alternative_methods}
\vspace{-6pt}
\end{figure*}

\noindent\textbf{Analysis of estimation error.}
Then we show that the estimation error of \methodabbr can be bounded by an infinitesimal, and thus the impact of \methodabbr on the estimation accuracy is insignificant.
The detailed proof is shown in \cref{subsec:bound-proof}.
\begin{theorem}
\vspace{-3pt}
\label{prop:bound}
Given the chosen $f_{\sT}(t)$,  when $t\in[0, \tilde{t})$, the difference between the optimal $\beps_\theta(\rvx, f_{\sT}(t))$ denoted as $\beps^*(\rvx, f_{\sT}(t))$, and $\beps(\rvx, t) = - \sigma_t \nabla_\rvx \log q_t(\rvx)$, can be bounded by
\begin{equation}
\left\Vert \beps^*\left(\rvx, f_{\sT}\left(t\right)\right) - \beps\left(\rvx, t\right)\right\Vert \leq \sigma_{\tilde{t}} K\left(\rvx\right) \Delta t + B\left(\rvx\right) \Delta \sigma_{\max},
\end{equation}
where
\begin{equation}
K\left(\rvx\right)=\sup_{t\neq \tau} \frac{\Vert \nabla_\rvx \log q_t\left(\rvx\right) - \nabla_\rvx \log q_\tau\left(\rvx\right)\Vert}{|t - \tau|}, \quad B\left(\rvx\right)=\sup_t \Vert \nabla_\rvx \log q_t\left(\rvx\right)\Vert,
\end{equation}
and $\Delta {\sigma_{\max}} = \max_{1\leq i\leq n}|\sigma_{t_i} - \sigma_{t_{i-1}}|$. Note that $K(\rvx)$ and $B(\rvx)$ are finite and 
$\lim_{\Delta t \rightarrow 0} \Delta \sigma_{\max} = 0$ for any continuous $\sigma_t$ where $\Delta t  \defeq \tilde{t} / n$, thus the difference converges to 0 as $\Delta t \rightarrow 0$.
Furthermore, the rate of convergence is at least $\frac{1}{2}$-order with respect to $\Delta t$.
\end{theorem}
The $\frac{1}{2}$-order convergence rate is relatively fast in optimization.
Given this bound, we think the introduced errors of \methodabbr are controllable.

\noindent\textbf{Reduction in Lipschitz constants.}
In \cref{fig:framework}b, we present the curve of $K(t, t^\prime)$ of \methodabbr on FFHQ $256\times 256$~\citep{karras2019style} (we provide results for continuous-time diffusion models and more results on other datasets in \cref{sec:app-results:lipschitz-constants}), showing that the Lipschitz constants $K(t, t^\prime)$ are significantly reduced by applying \methodabbr.

\noindent\textbf{Improvement in stability.}
To further verify the stability of \methodabbr, we evaluate the impact of a small perturbation added to the input. 
Specifically, we add a small noise with a growing scale to the $\rvx_{\tilde{t}}$, where $\tilde{t}$ is set to a default value of 100, and observe the resulting difference in the predicted value of $\rvx_0$, for both \methodabbr and baseline.
Our results, as shown in \cref{fig:stability}, illustrate that \methodabbr exhibits better stability than the baseline, as its predictions are less affected by perturbations. 

\noindent\textbf{Comparison with some alternative methods.}
Although achieving impressive performance as detailed in \cref{sec:exp}, \methodabbr introduces no modifications to the network architecture or loss function, thereby not incurring any additional computational cost.
\textbf{1) Regularization}: In contrast, an alternative potential approach is imposing restrictions on the Lipschitz constants via regularization techniques.
It necessitates the computation of $\frac{\partial \beps_\theta(\rvx, t)}{\partial t}$, consequently diminishing training efficiency.
\textbf{2) Modification of noise schedules}: Furthermore, \methodabbr preserves the forward process unaltered.
Conversely, another potential method involves the modification of noise schedules.
Recall that the issue of Lipschitz singularities only arises when the noise schedule satisfies $\frac{d\alpha_t}{dt}|_{t=0} \neq 0$. 
Therefore, it becomes feasible to adjust the noise schedule to meet the requirement $\frac{d\alpha_t}{dt}|_{t=0} = 0$, thus mitigating the problem of Lipschitz singularities.
The detailed methods for modifying noise schedules are provided in \cref{sec:alternative-methods:noise-schedules}.
Although this modification seems feasible, it results in tiny amounts of noise at the beginning stages of the diffusion process, leading to inaccurate predictions.
\textbf{3) Remap}: In addition, remap is another possible method, which designs a remap function $\lambda=f(t)$ as the conditional input of the network, namely, $\beps_\theta(\rvx, f(t))$.
By carefully designing $\lambda=f(t)$, it can significantly stretch the interval with large Lipschitz constants.
For example, $f(t) = 1/t$ and $f^{-1}(\lambda) = \text{sigmoid}(\lambda)$ are two simple choices.
In this way, Remap can efficiently reduce the Lipschitz constants regarding the conditional inputs of the network, $\frac{\partial \beps_\theta(\rvx, t)}{\partial \lambda}$.
However, since we uniformly sample $t$ both in training and inference, what should be focused on is the Lipschitz constants regarding $t$, $\frac{\partial \beps_\theta(\rvx, t)}{\partial t}$, which can not be influenced by remap.
We also consider the situation of uniformly sampling $\lambda$, which can significantly hurt the quality of generated images.
We show the quantitative evaluation in \cref{fig:alternative_methods} and put the detailed analysis in \cref{sec:alternative-methods:remap}.
Empirically, \methodabbr surpasses not only the baseline but also all of these alternative methods, where the results are demonstrated in \cref{fig:alternative_methods}.
For a more in-depth discussions, please refer to Section \ref{sec:alternative-methods}.

\begin{figure*}[t]
\hspace{1pt}
\begin{minipage}[c]{0.45\textwidth}
\centering
% \vspace{3pt}

\includegraphics[width=1.0\linewidth]{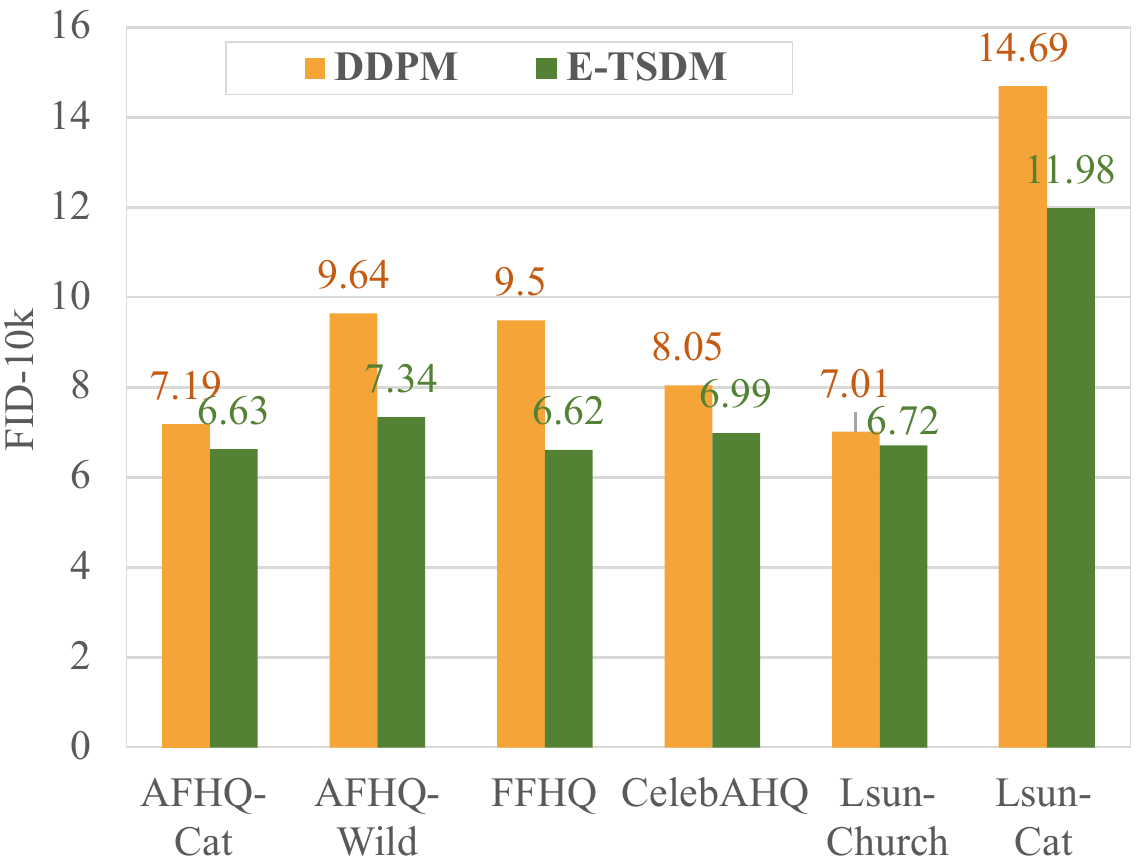}
\centering

\vspace{-5pt}
\caption{\textbf{Quantitative comparison} on various datasets with $256\times256$ resolution.
        All experiments are evaluated with FID-10k $\downarrow$.}
\label{fig:performance}
\end{minipage}
\hfill
\begin{minipage}[c]{0.48\textwidth}
\vspace{5pt}
\centering
\captionof{table}{\textbf{Quantitative comparison} on FFHQ \citep{karras2019style}. $\ast$ denotes our reproduced result with the same network as \methodabbr-large.
}
\label{tab:sota}
% \vspace{8pt}
\setlength{\tabcolsep}{3pt}
% \small
\centering
  \begin{tabular}{lc}
    \toprule
    Model    & FID-50k $\downarrow$    \\
    \midrule
    StyleGAN2$+$ADA$+$bCR                              & \multirow{2}{*}{3.62} \\
    ~\citep{karras2020training}                         &     \\
    \midrule
    Soft-Truncation~\citep{kim2021soft}                 & 5.54 \\ 
    P2-DM~\citep{choi2022perception}                    & 6.92 \\ 
    LDM~\citep{rombach2022high}                         & 4.98  \\
    DDPM~\citep{ho2020denoising}                        & ~~6.88$^*$  \\
    \methodabbr (ours)                                 & 5.21  \\
    \methodabbr-large (ours)                           & $\bf4.22$  \\
    \bottomrule
  \end{tabular}
% \vspace{-5pt}
\end{minipage}
\hspace{1pt}
\vspace{-20pt}
\end{figure*}

\section{Experiments}\label{sec:exp}

In this section, we present compelling evidence that our \methodabbr outperforms existing approaches on a variety of datasets.
To achieve this, we first detail the experimental setup used in our studies in \cref{subsec:settings}. 
Subsequently, in \cref{subsec:performance}, we compare the synthesis performance of \methodabbr against that of the baseline on various datasets.
In \cref{subsec:ablation}, we conduct multiple ablation studies and quantitative analysis from two perspectives.
Firstly, we demonstrate the generalizability of \methodabbr by implementing it on continuous-time diffusion models and varying the noise schedules. 
Secondly, we investigate the impact of varying the number of conditions $n$ in $t\in[0,\tilde{t})$ and the length of the interval $\tilde{t}$, which are important hyperparameters.
Moreover, we demonstrate in \cref{subsec:acceleration} that our method can be effectively combined with popular fast sampling techniques.
Finally, we show that \methodabbr can be applied to conditional generation tasks, such as super-resolution, in \cref{subsec:conditional-generation}. 

\subsection{Experimental setup}\label{subsec:settings}

\noindent\textbf{Implementation details.}
All of our experiments utilize the settings of DDPM~\citep{ho2020denoising} (see more details in \cref{subsec:app-hyper}).
Besides, we utilize a more developed structure of unet~\citep{dhariwal2021diffusion} than that of DDPM~\citep{ho2020denoising} for all experiments containing reproduced baseline.
Given that the model size is kept constant, the speed and memory requirements for training and inference for both the baseline and \methodabbr are the same. 
Except for the ablation studies in \cref{subsec:ablation}, all other experiments fix $\tilde{t}=100$ for \methodabbr and use five conditions ($n=5$) in the interval $t\in[0,\tilde{t})$, which we have found to be a relatively good choice in practice.
Furthermore, all experiments are trained on NVIDIA A100 GPUs.
\noindent\textbf{Datasets.}
We implement \methodabbr on several widely evaluated datasets containing FFHQ $256\times 256$~\citep{karras2019style}, CelebAHQ $256\times 256$~\citep{karras2017progressive}, AFHQ-Cat $256\times 256$, AFHQ-Wild $256\times 256$~\citep{choi2020stargan}, Lsun-Church $256\times 256$ and Lsun-Cat $256\times 256$~\citep{yu2015lsun}.
\noindent\textbf{Evaluation metrics.}
To assess the sampling quality of \methodabbr, we utilize the widely-adopted Frechet inception distance (FID) metric~\citep{heusel2017gans}. 
Additionally, we use the peak signal-to-noise ratio (PSNR) to evaluate the performance of the super-resolution task.

\begin{table}
\renewcommand{\arraystretch}{0.75}
  \caption{
    \textbf{Quantitative comparison} between DDPM baseline~\citep{ho2020denoising} and our proposed \methodabbr on both discrete-time and continuous-time scenarios with different noise schedules, on FFHQ $256\times 256$~\citep{karras2019style} using FID-10k $\downarrow$ as the evaluation metric. 
    Experimental results illustrate that \textbf{\methodabbr can be generalized to other noise schedules and is still effective in the context of continuous-time diffusion models}.
  }
  \vspace{-6pt}
  \label{tab:ablation-1}
  \centering
  \setlength{\tabcolsep}{10pt}
  \resizebox{\textwidth}{!}{
  \begin{tabular}{c|c|c|c|c|c|c}
       \toprule
       \multirow{3}{*}{Settings}   & \multirow{3}{*}{Method}    & \multicolumn{5}{c}{Noise schedule}  \\
       \cmidrule{3-7}
       &    & Linear  & Quadratic  & Cosine  & Cosine-shift  & Zero-terminal-SNR  \\
       \midrule
       \multirow{2}{*}{Discrete}  & Baseline  & 9.50  & 13.79  & 27.17  & 14.51  & 11.66  \\
       & \methodabbr & $\bf6.62$  & $\bf9.69$  & $\bf26.08$  & $\bf11.20$  & $\bf8.58$  \\
       \midrule
       \multirow{2}{*}{Continuous}  & Baseline  & 9.53  & 14.26  & 25.65  & 12.80  & 10.89  \\
       & \methodabbr & $\bf6.95$  & $\bf9.66$  & $\bf16.80$  & $\bf9.94$  & $\bf8.96$  \\
       \bottomrule
  \end{tabular}}
  \vspace{-12pt}
\end{table}

\subsection{Synthesis performance}\label{subsec:performance}

We have demonstrated that \methodabbr can effectively mitigate the large Lipschitz constants near $t=0$ in \cref{fig:framework} b, as detailed in \cref{sec:method}.
In this section, we conduct a comprehensive comparison between \methodabbr and DDPM baseline~\citep{ho2020denoising} on various datasets to show that \methodabbr can improve the synthesis performance.
The quantitative comparison is presented in \cref{fig:performance}, which clearly illustrates that \methodabbr outperforms the baseline on all evaluated datasets.
Furthermore, as depicted in \cref{app:more-samples}, the samples generated by \methodabbr on various datasets demonstrate its ability to generate high-fidelity images.
Remarkably, to the best of our knowledge, as shown in \cref{tab:sota}, we set a new state-of-the-art benchmark for diffusion models on FFHQ $256\times256$~\citep{karras2019style} using a large version of our approach (see details in \cref{subsec:app-hyper}).

\subsection{Quantitative analysis}\label{subsec:ablation}

In this section, we demonstrate the generalizability of \methodabbr by implementing it on continuous-time diffusion models and varying the noise schedules.
In addition, to gain a deeper understanding of the properties of \methodabbr, we investigate the critical hyperparameters of \methodabbr by varying the length of the interval $\tilde{t}$ to share the timestep conditions, and the number of sub-intervals $n$.

\subsubsection{Quantitative analysis on the generalizability of \methodabbr}\label{ssubsec:para-schedule}

To ensure the generalizability of \methodabbr beyond specific settings of DDPM~\citep{ho2020denoising}, we conduct a thorough investigation of \methodabbr on other popular noise schedules, as well as implement a continuous-time version of \methodabbr.
Specifically, we explore the three popular ones including linear, quadratic and cosine schedules~\citep{nichol2021improved}, and two newly proposed ones, which are cosine-shift~\citep{hoogeboom2023simple} and zero-terminal-SNR~\citep{lin2023common} schedules.

As shown in \cref{tab:ablation-1}, our \methodabbr achieves excellent performance across different noise schedules.
Besides, the comparison of Lipschitz constants between \methodabbr and baseline on different noise schedules, as illustrated in \cref{sec:app-results:lipschitz-constants}, show that \methodabbr can mitigate the Lipschitz singularities issue besides the scenario of the linear schedule, highlighting that its effects are independent of the specific noise schedule.
Additionally, the continuous-time version of \methodabbr outperforms the corresponding baseline, indicating that \methodabbr is effective for both continuous-time and discrete-time diffusion models.
We provide the curves of the Lipschitz constants $K(t, t^\prime)$ in \cref{fig:lipschitz-continuous} to compare continuous-time \methodabbr with its baseline on the linear schedule, showing that \methodabbr can mitigate Lipschitz singularities in the continuous-time scenario.

\begin{figure*}[t]
\centering
\begin{overpic}[width=0.99\linewidth]{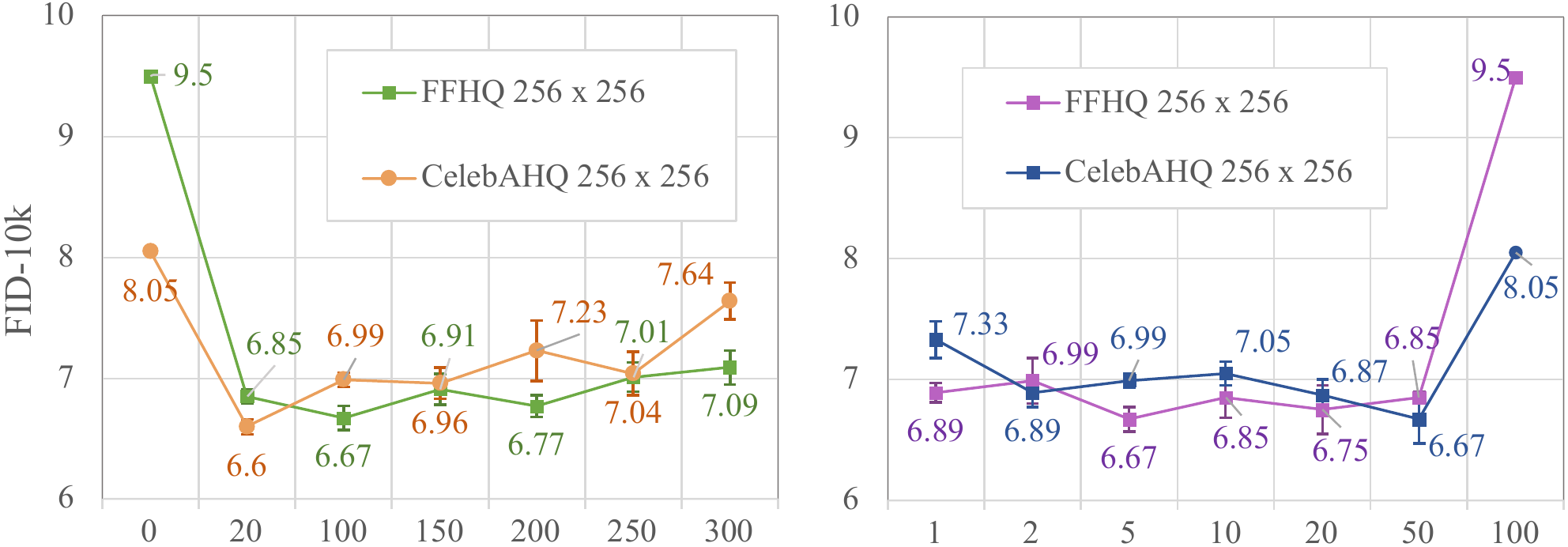}
 \put(17,-3){(a) Interval Length $\tilde{t}$}
 \put(67,-3){(b) Timestep Number $n$}

\end{overpic}
\vspace{4pt}
\caption{ \textbf{Ablation study} on the length of the interval $t\in [0, \tilde{t})$ to share the timestep conditions, $\tilde{t}$, and the number of sub-intervals in this interval, $n$, using FID-10k $\downarrow$ as the evaluation metric.
We repeat each experiment three times and provide the error bars.}
\vspace{-6pt}
\label{fig:ablation}
\end{figure*}

\begin{table}
\renewcommand{\arraystretch}{0.85}
  \caption{
    \textbf{Quantitative comparison} on FFHQ $256\times 256$~\citep{karras2019style} between DDPM~\citep{ho2020denoising} and our proposed \methodabbr utilizing different fast samplers, DDIM~\citep{song2020denoising} and DPM-Solver~\citep{lu2022dpm}, varying the number of function evalutaions (NFE). 
    FID-10k$\downarrow$ is used as the evaluation metric, and DPM-Solver-$k$ represents the $k$-th-order DPM-Solver.
    Results indicate that \textbf{our approach well supports the popular fast samplers}.
    }
    \vspace{-6pt}
  \label{tab:acceleration}
  \centering
  \setlength{\tabcolsep}{10pt}
  \begin{tabular}{c|c|c|c|c|c|c|c}
    \toprule
    \multicolumn{2}{c|}{Fast Samplers}    &\multicolumn{2}{c|}{DPM-Solver-3}  &\multicolumn{2}{c|}{DPM-Solver-2}  &\multicolumn{2}{c}{DDIM} \\
    \midrule
    \multicolumn{2}{c|}{NFE}   &20   &50   &20   &50   &50   &200 \\
    \midrule
    \multirow{2}{*}{Method}  &DDPM  &21.91  &24.48  &22.21  &24.80  &21.80  &23.16 \\  
    &\methodabbr  &$\bf16.97$  &$\bf13.52$  &$\bf17.28$  &$\bf14.14$  &$\bf19.34$  &$\bf13.71$ \\
    \bottomrule
  \end{tabular}
  \vspace{-12pt}
\end{table}

\subsubsection{Quantitative analysis on $n$ and $\tilde{t}$}\label{ssubsec:ablation}

\methodabbr involves dividing the target interval $t \in [0, \tilde{t})$ with large Lipschitz constants into $n$ sub-intervals and sharing timestep conditions within each sub-interval.
Accordingly, the choices of $\tilde{t}$ and $n$ have significant impacts on the performance of \methodabbr. 
Intuitively, $\tilde{t}$ should be a relatively small value, therefore representing an interval near zero point.
As for $n$, it should not be too large or too small.
If $n$ is too small, it forces the network to adapt to too many noise levels with a single timestep condition, thus leading to inaccuracy.
Conversely, if the value of $n$ is set too large, the reduction of Lipschitz constants is insufficient, where the extreme situation is baseline.

In this section, we meticulously assess the impacts of $\tilde{t}$ and $n$ on various datasets. 
We present the outcomes on FFHQ $256\times256$~\citep{karras2019style} and CelebAHQ $256\times 256$~\citep{karras2017progressive} for each hyperparameter in \cref{fig:ablation}, while leaving the remaining results in \cref{sec:app-results:quantitative and ablation:ablation}.
Specifically, in the experiments of $\tilde{t}$, we maintain the length of each sub-interval, namely, $\tilde{t} / n$, unchanged, while in the experiments of $n$, we maintain the $\tilde{t}$ unchanged.
The results for $\tilde{t}$ in \cref{fig:ablation} a demonstrate that \methodabbr performs well when $\tilde{t}$ is relatively small.
However, as $\tilde{t}$ increases, the performance of \methodabbr deteriorates gradually.
Furthermore, the results for $n$ are shown in \cref{fig:ablation} b, from which we observe a rise in FID when $n$ was too small, for instance, when $n=2$.
Conversely, when $n$ is too large, such as $n=100$, the performance deteriorates significantly.
Although \methodabbr performs well for most $n$ and $\tilde{t}$ values, considering the results on all of the evaluated datasets (see remaining results in \cref{sec:app-results:quantitative and ablation:ablation}), $n=5$ and $\tilde{t}=100$ are recommended to be good choices to avoid cumbersome searches or a good starting point for further exploration when applying \methodabbr.

\subsection{Fast sampling}\label{subsec:acceleration}

With the development of fast sampling algorithms, it is crucial that \methodabbr can be effectively combined with classic fast samplers, such as DDIM~\citep{song2020denoising} and DPM-Solver~\citep{lu2022dpm}.
To this end, we incorporate both DDIM~\citep{song2020denoising} and DPM-Solver~\citep{lu2022dpm} into \methodabbr for fast sampling in this section.
It is worth noting that the presence of large Lipschitz constants can have a more detrimental impact on the efficiency of fast sampling compared to full-timestep sampling, as numerical solvers typically depend on the similarity between function values and their derivatives on adjacent steps.
When using fast sampling algorithms with larger discretization steps, it becomes necessary for the functions to exhibit better smoothness, which in turn corresponds to smaller Lipschitz constants.
Hence, it is anticipated that the utilization of \methodabbr will lead to an improvement in the generation performance of fast sampling methods.

As presented in \cref{tab:acceleration}, we observe that \methodabbr significantly outperforms the baseline when using the same number of function evaluations (NFE) for fast sampling, which is under expectation.
Besides, the advantage of \methodabbr becomes more pronounced when using higher order sampler (from DDIM to DPM-Solver), indicating better smoothness when compared to the baseline.
Notably, for both DDIM and DPM-Solver, we observe an abnormal phenomenon for baseline, whereby the performance deteriorates as NFE increases.
This phenomenon has been previously noted by several works~\citep{lu2022dpm, lu2022dpmplus, li2023era}, but remains unexplained.
Given that this phenomenon is not observed in \methodabbr, we hypothesize that it may be related to the improvement of smoothness of the learned network.
We leave further verification of this hypothesis for future work.

\subsection{Conditional generation}\label{subsec:conditional-generation}

In order to explore the potential for extending \methodabbr to conditional generation tasks, we further investigate its performance in the super-resolution task, which is one of the most popular conditional generation tasks.
Specifically, we test \methodabbr on the FFHQ $256\times 256$ dataset, using the $64\times64 \rightarrow 256\times256$ pixel resolution as our experimental settings.
For the baseline in the super-resolution task, we utilize the same network structure and hyper-parameters as those employed in the baseline presented in \cref{subsec:settings}, but incorporate a low-resolution image as an additional input.
Besides, for \methodabbr, we adopt a general setting with $n=5$ and $\tilde{t}=100$.
As illustrated in \cref{fig:super-resolution}, we observe that the baseline tends to exhibit a color bias compared to real images, which is mitigated by \methodabbr.
Quantitatively, our results indicate that \methodabbr outperforms the baseline on the test set, achieving an improvement in PSNR from 24.64 to 25.61. These findings suggest that \methodabbr holds considerable promise for application in conditional generation tasks.
\section{Conclusion}\label{sec:conclusion}

In this paper, we elaborate on the infinite Lipschitz of the diffusion model near the zero point from both theoretical and empirical perspectives, which hurts the stability and accuracy of the diffusion process.
A novel \methodabbr is further proposed to mitigate the corresponding singularities in a timestep-sharing manner.
Experimental results demonstrate the superiority of our method in both performance and adaptability to the baselines, including unconditional generation, conditional generation, and fast sampling. 
This paper may not only improve the performance of diffusion models, but also help to make up the critical research gap in the understanding of the rationality underlying the diffusion process.

\noindent\textbf{Limitations.}
Although \methodabbr has demonstrated significant improvements in various applications, it has yet to be verified on large-scale text-to-image generative models.
As \methodabbr reduces the large Lipschitz constants by sharing conditions, it is possible to lead to a decrease in the effectiveness of large-scale generative models. 
Additionally, the reduction of Lipschitz constants to zero within each sub-interval in \methodabbr may introduce unknown and potentially harmful effects.
\section*{Acknowledgments}\label{sec:acknowledgments}

We would like to thank the four anonymous reviewers for spending time and effort and bringing in constructive questions and suggestions, which helped us greatly to improve the quality of the paper.
We would like to also thank the Program Chairs and Area Chairs for handling this paper and providing valuable and comprehensive comments.
In addition, this research was funded by the Alibaba Innovative Research (AIR) project.

{
\bibliographystyle{iclr2024_conference}
\bibliography{ref}
}

\clearpage
\appendix
\newcommand{\AppendixPrefix}{A}
\renewcommand{\thefigure}{\AppendixPrefix\arabic{figure}}
\setcounter{figure}{0}
\renewcommand{\thetable}{\AppendixPrefix\arabic{table}} 
\renewcommand{\thealgorithm}{\AppendixPrefix\arabic{algorithm}} 
\setcounter{algorithm}{0}
\setcounter{table}{0}
\renewcommand{\theequation}{\AppendixPrefix\arabic{equation}} 
\setcounter{equation}{0}
\section*{Appendix}

\section{Overview}

This supplementary material is organized as follows.
First, to facilitate the reproducibility of our experiments, we present implementation details, including hyper-parameters in \cref{subsec:app-hyper} and algorithmic details in \cref{subsec:app-code}.
Next, in \cref{sec:app-derivation}, we provide all details of deduction involved in the main paper.
Finally, we present additional experimental results in support of the effectiveness of \methodabbr.

\section{Implementation details}\label{sec:app-details}

\subsection{Hyper-parameters}\label{subsec:app-hyper}

The hyper-parameters used in our experiments are shown in \cref{tab:app-hyperparameter}, and we use identical hyper-parameters for all evaluated datasets for both \methodabbr and their corresponding baselines.
Specifically, we follow the hyper-parameters of DDPM~\citep{ho2020denoising} but adopt a more advanced structure of U-Net~\citep{dhariwal2021diffusion} with residual blocks from BigGAN~\citep{brock2018large}.
The network employs a block consisting of fully connected layers to encode the timestep, where the dimensionality of hidden layers for this block is determined by the timestep channels shown in \cref{tab:app-hyperparameter}.
\begin{table*}[!ht]
\caption{
    \textbf{Hyper-parameters} of \methodabbr and our reproduced baseline.
}
\label{tab:app-hyperparameter}
% \vspace{-15pt}
\begin{center}
\setlength{\tabcolsep}{4pt}
\begin{tabular}{lcc}
\toprule
& Normal version & Large version  \\
\midrule 
$T$                   & 1,000            & 1,000           \\
$\beta_t$             & linear           & linear          \\
Model size            & 131M             & 692M            \\
Base channels         & 128              & 128             \\
Channels multiple     & (1,1,2,2,4,4)    & (1,1,2,4,6,8)   \\
Heads channels        & 64               & 64              \\
Self attention        & 32,16,8          & 32,64,8         \\
Timestep channels     & 512              & 2048            \\
BigGAN block          & \ding{51}        & \ding{51}       \\
Dropout               & 0.0              & 0.0             \\
Learning rate         & 1e$^{-4}$        & 1e$^{-4}$       \\
Batch size            & 96               & 64              \\
Res blocks            & 2                & 4               \\
EMA                   & 0.9999           & 0.9999          \\
Warmup steps          & 0                & 0               \\
Gradient clip         & \ding{55}        & \ding{55}       \\
\bottomrule
\end{tabular}
\end{center}
\vspace{-10pt}
\end{table*}
Moreover, we scale up the network to achieve the state-of-the-art results of diffusion models on FFHQ $256\times 256$~\citep{karras2019style}, and therefore we provide the hyper-parameters of the large version of \methodabbr in \cref{tab:app-hyperparameter}.

\subsection{Algorithm details}\label{subsec:app-code}

In this section, we provide a detailed description of the \methodabbr algorithm, including the training and inference procedures as shown in \cref{alg:training} and \cref{alg:sampling}, respectively.
Our method is simple to implement and requires only a few steps. 
Firstly, a suitable length of the interval $\tilde{t}$ should be selected for sharing conditions, along with the corresponding number of timestep conditions $n$ in the target interval $t \in [0, \tilde{t})$.
While performing a thorough search for different datasets can achieve better performance, the default settings $\tilde{t}=100$ and $n=5$ are recommended when \methodabbr is applied without a thorough search.
\begin{algorithm}[!ht]
    \renewcommand{\algorithmicensure}{\textbf{Output:}}
    \caption{Training of \methodabbr}
    \label{alg:training}
    \begin{algorithmic}[1]
        \Require The length of the target interval $\tilde{t}$.
        \Require The number of conditions $n$.
        \Require Model $\rvepsilon_\theta$ to be trained.
        \Require Data set $\mathcal{D}$.

        \State Uniformly divide the target interval $t\in[0, \tilde{t})$ into $n$ sub-intervals to get the corresponding timestep partition schedule $\sT = \{t_0, t_1, \dots, t_n\}$.

        \Repeat
            \State $\rvx_0 \sim \mathcal{D}$
            \State $t \sim \text{Uniform}(\{1, \dots, T\})$
            \If{$t < \tilde{t}$}
                \State $\hat{t}=\max_{1\le i\le n}\{t_{i-1}\in \sT: t_{i-1} \le t\}$
            \Else
                \State $\hat{t}=t$
            \EndIf
            \State $\beps \sim \mathcal{N}(\mathbf{0}, \mathbf{I})$
            \State Take gradient descent step on \\ 
            \quad \quad \quad $\nabla_\theta \Vert\rvepsilon - \rvepsilon_\theta (\alpha_t\rvx_0 + \sigma_t \rvepsilon, \hat{t}) \Vert^2$
        \Until{converged}
    \end{algorithmic}
\end{algorithm}

\begin{algorithm}[!ht]
    \renewcommand{\algorithmicensure}{\textbf{Output:}}
    \caption{Sampling of \methodabbr}
    \label{alg:sampling}
    \begin{algorithmic}[1]
        \Require The length of the target interval $\tilde{t}$.
        \Require The number of conditions $n$.
        \Require A trained model $\rvepsilon_\theta$.

        \State Uniformly divide the target interval $t\in[0, \tilde{t})$ into $n$ sub-intervals to get the corresponding timestep partition schedule $\sT = \{t_0, t_1, \dots, t_n\}$.

        \State $\rvx_T \sim \mathcal{N}(\mathbf{0},\mathbf{I})$
        \For{$t=T,\dots, 1$}
            \If{$t < \tilde{t}$}
                \State $\hat{t}=\max_{1\le i\le n}\{t_{i-1}\in \sT: t_{i-1} \le t\}$
            \Else
                \State $\hat{t}=t$
            \EndIf
            \If{$t>1$}
                \State $\rvz \sim \mathcal{N}(\mathbf{0}, \mathbf{I})$ 
            \Else
                \State $\rvz = 0$
            \EndIf
        \State $\rvx_{t-1}=\frac{\alpha_{t-1}}{\alpha_t} \left(\rvx_t-\frac{\beta_t}{\sigma_t} \rvepsilon_\theta(\rvx_t, \hat{t})\right) + \eta_t \rvz$
        \EndFor \\
        \Return{$\rvx_0$}
    \end{algorithmic}
\end{algorithm}
Next, the target interval $t \in [0, \tilde{t})$ should be divided into $n$ sub-intervals, and the boundaries for each sub-interval should be calculated to generate the partition schedule $\sT = \{t_0, t_1, \dots, t_n\}$. 
Finally, during both training and sampling, the corresponding left boundary $\hat{t}$ for each timestep in the target interval $t \in [0, \tilde{t})$ should be determined according to $\sT$, and used as the conditional input of the network instead of $t$.

\begin{figure*}[t]
\centering
\begin{overpic}[width=0.45\linewidth]{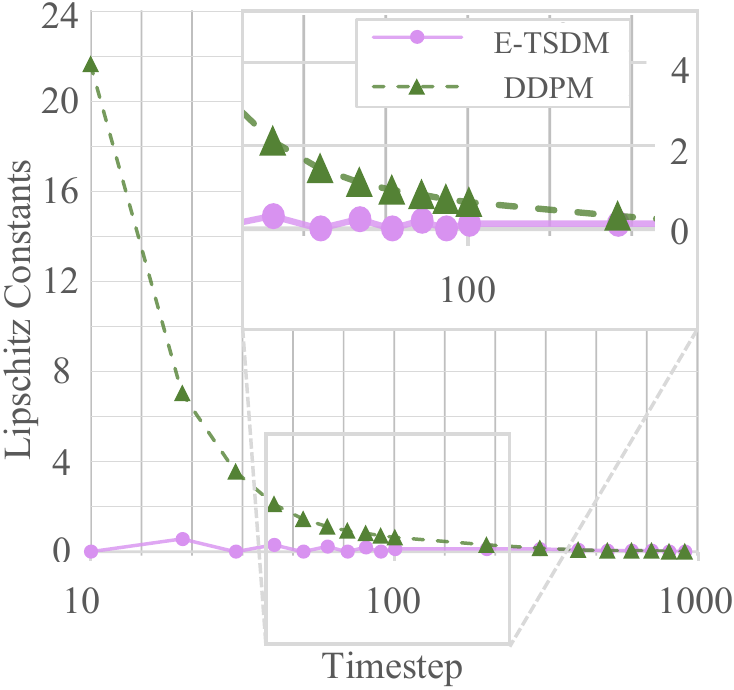}
\end{overpic}
\vspace{-8pt}
\caption{\textbf{Quantitative comparison} of the Lipschitz constants between \textbf{continuous-time} \methodabbr and \textbf{continuous-time} DDPM~\citep{ho2020denoising}.
Results show that \methodabbr can efficiently reduce the Lipschitz constants in \textbf{continuous-time} scenarios.
}
\label{fig:lipschitz-continuous}
\vspace{-10pt}
\end{figure*}

\section{Derivation of formulas}\label{sec:app-derivation}

In this section, we provide detailed derivations as a supplement to the main paper.
The derivations are divided into three parts, firstly we prove that the key assumption of the occurrence of Lipschitz singularities, $\left. \frac{\dd\alpha_t}{\dd t} \right\vert_{t=0} \neq 0$, holds for mainstream noise schedules including linear, quadratic, and cosine schedules.
Therefore, all of the diffusion models utilizing these noise schedules suffer from the issue of Lipschitz singularities.
Then we show that Lipschitz singularities also plague the v-prediction~\citep{salimans2022progressive} models. 
Considering that most of the diffusion models are noise-prediction or v-prediction models, the Lipschitz singularities problem is an important issue for the community of diffusion models.
Finally, we demonstrate the detailed derivation of \cref{prop:bound}, showing that the errors introduced by \methodabbr can be bounded by an infinitesimal and thus are insignificant.

\begin{figure*}[t]
\begin{center}
\begin{overpic}[width=0.99\linewidth]{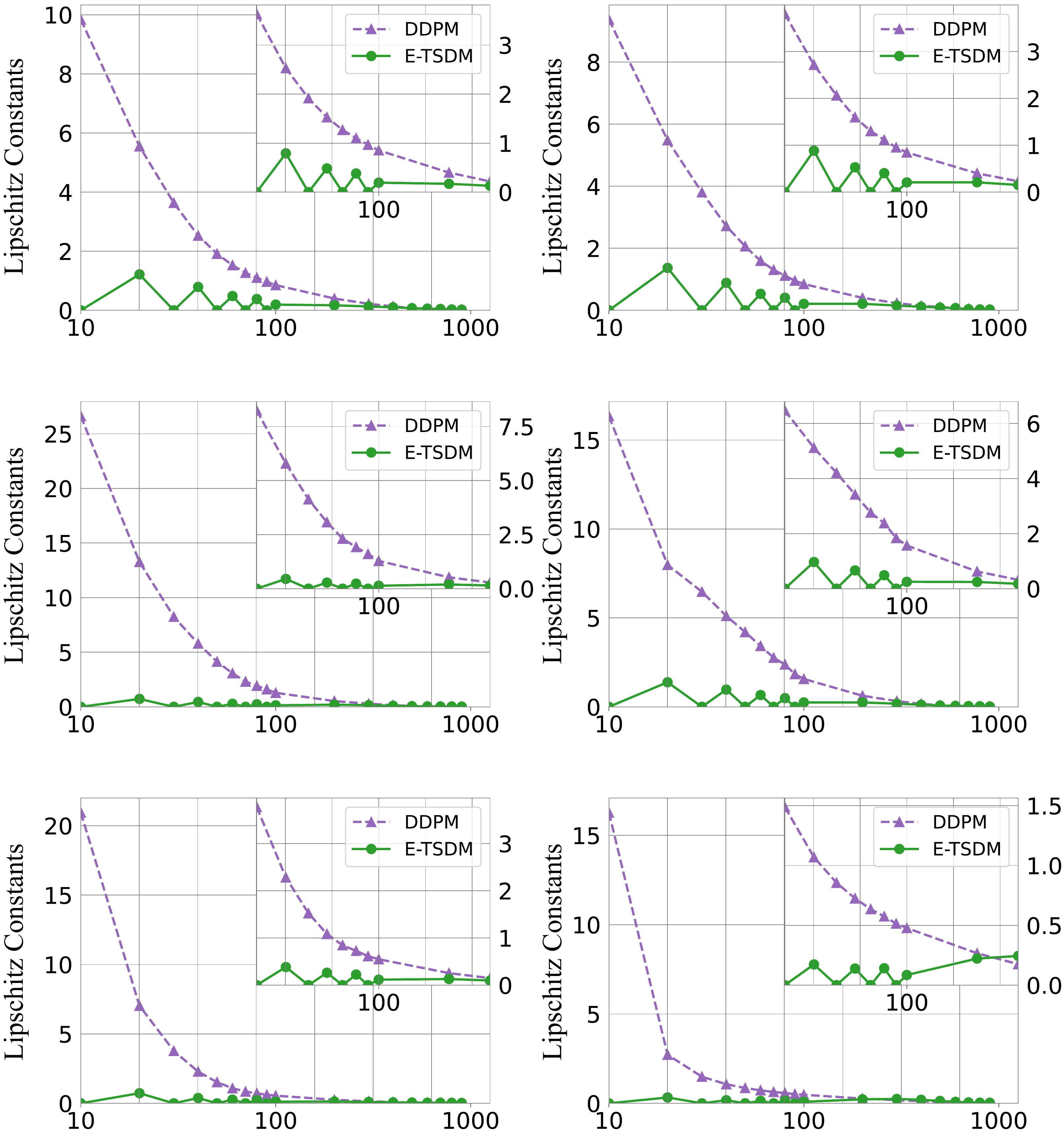}
  \put(14,68){(a) AFHQ-Cat $256\times256$}

  \put(58,68){(b) AFHQ-Wild $256\times256$}
  
  \put(14,32){(c) Lsun-Cat $256\times256$}

  \put(57.5,32){(d) Lsun-Church $256\times256$}

  \put(13.5,-3){(e) CelebAHQ $256\times256$}

  \put(58,-3){(f) FFHQ $256\times256$ using}
  \put(62,-5.5){quadratic schedule}
\end{overpic}
\end{center}
\vspace{12pt}
\caption{\textbf{Quantitative comparison} of Lipschitz constants between \methodabbr and DDPM baseline~\citep{ho2020denoising} \textbf{on various datasets}, including (a) AFHQ-Cat~\citep{choi2020stargan}, (b) AFHQ-Wild~\citep{choi2020stargan}, (c) Lsun-Cat $256\times256$~\citep{karras2019style}, (d) Lsun-Church $256\times256$~\citep{karras2019style}, and (e) CelebAHQ $256\times256$~\citep{karras2017progressive} using the linear schedule. (f) \textbf{Quantitative comparison} of Lipschitz constants between \methodabbr and DDPM baseline~\citep{ho2020denoising} on FFHQ $256\times256$~\citep{karras2019style} using the \textbf{quadratic schedule}.
}
\label{fig:supp-lipschitz-constants}
\vspace{-20pt}
\end{figure*}

\subsection{$\dd \alpha_t / \dd t$ for widely used noise schedules at zero point}\label{subsec:alpha-derivatives}

\begin{figure*}[t]
\begin{center}
\begin{overpic}[width=0.99\linewidth]{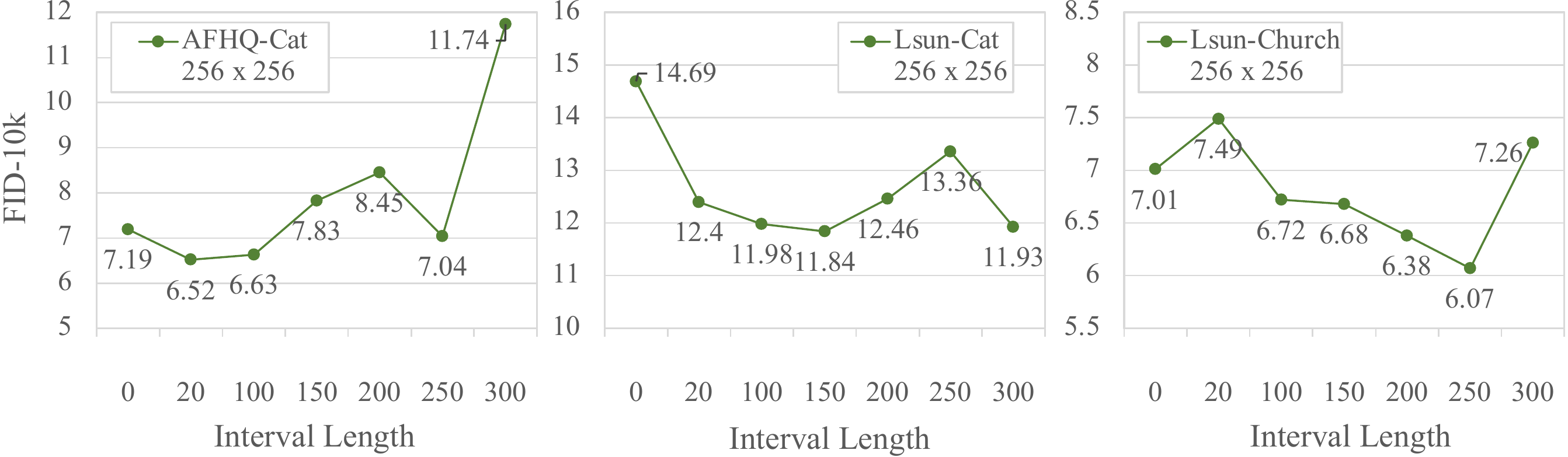}

  \put(7,-3){(a) AFHQ-Cat $256\times256$}
  % \put(14,-2.5){Interval Length}
  % \put(53,52){\rotatebox{90}{FID-10k}}

  \put(40,-3){(b) LSUN-Cat $256\times256$}
  % \put(46,-2.5){Interval Length}
  % % \put(2,11){\rotatebox{90}{FID-10k}}

  \put(70,-3){(c) LSUN-Church $256\times256$}
  % \put(78,-2.5){Interval Length}
\end{overpic}
\end{center}
\vspace{4pt}
\caption{\textbf{Ablation study} on the length of the interval $t\in [0, \tilde{t})$ to share the timestep conditions, $\tilde{t}$, using FID-10k $\downarrow$ as the evaluation metric.
}
\label{fig:supp-ab-t}
% \vspace{-5pt}
\end{figure*}

\begin{figure*}[t]
\begin{center}
% \fbox{\rule[-.5cm]{0cm}{12cm} \rule[-.5cm]{16cm}{0cm}}
% \includegraphics[width=1\textwidth]{figures/lipschitz constants.pdf}
\begin{overpic}[width=0.99\linewidth]{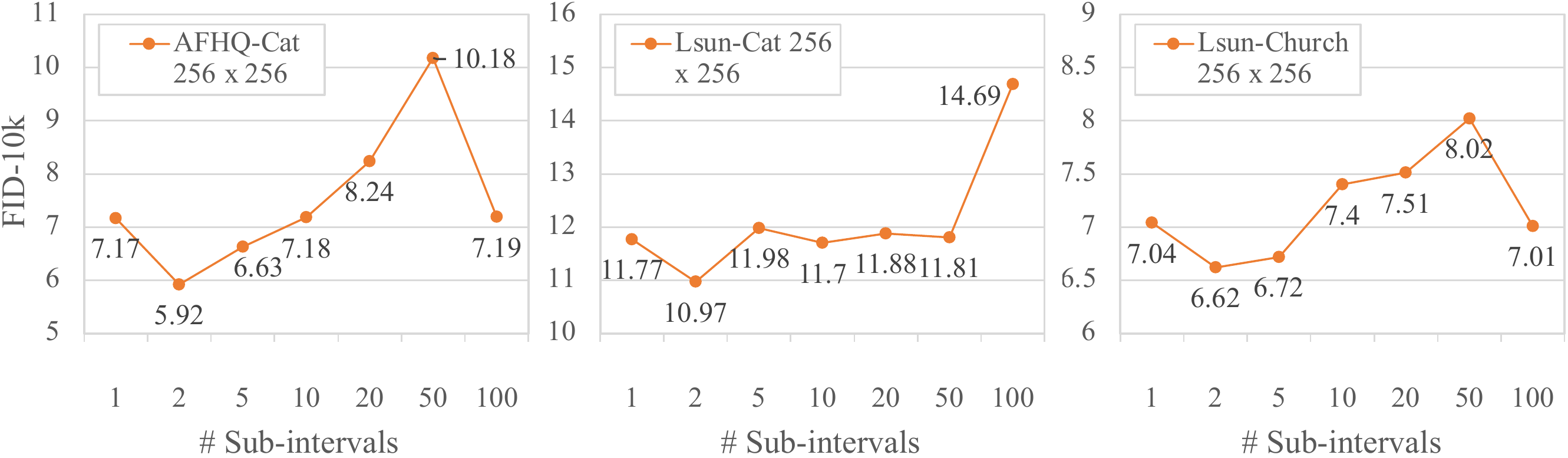}

  \put(7,-3){(a) AFHQ-Cat $256\times256$}
  % \put(14,-2.5){\#Sub-intervals}
  % \put(53,52){\rotatebox{90}{FID-10k}}

  \put(40,-3){(b) LSUN-Cat $256\times256$}
  % \put(46,-2.5){\#Sub-intervals}
  % % \put(2,11){\rotatebox{90}{FID-10k}}

  \put(70,-3){(c) LSUN-Church $256\times256$}
  % \put(78,-2.5){\#Sub-intervals}
  % \put(53,11){\rotatebox{90}{FID-10k}}
\end{overpic}
\end{center}
\vspace{4pt}
\caption{\textbf{Ablation study} on the number of sub-intervals in this interval, $n$, using FID-10k $\downarrow$ as the evaluation metric.
}
\label{fig:supp-ab-n}
% \vspace{-5pt}
\end{figure*}

We have already shown that for an arbitrary complex distribution, given a noise schedule, if $\left. \frac{\dd\alpha_t}{\dd t} \right\vert_{t=0} \neq 0$, then we often have $\lim \sup_{t\rightarrow 0+} \big\lVert \frac{\partial \beps_\theta(\rvx, t)}{\partial t} \big\rVert \rightarrow \infty$, indicating the infinite Lipschitz constants around $t=0$.
In this section, we prove that $\left. \frac{\dd\alpha_t}{\dd t}\right\vert_{t=0} \neq 0$ stands for three mainstream noise schedules including linear schedule, quadratic schedule and cosine schedule.
 
\subsubsection{$\dd \alpha_t / \dd t$ for linear and quadratic schedules at zero point}\label{ssubsec:alpha-derivatives-lq}

Linear and quadratic schedules are first proposed by \citet{ho2020denoising}.
Both of them determine $\{\alpha_t\}_{t=1}^T$ by a pre-designed positive sequence $\{\beta_t\}_{t=1}^T$ and the relationship $\alpha_t := \prod_{i=1}^t \sqrt{1 - \beta_i}$.
Note that $t\in \{1,2,\cdots, T\}$ is a discrete index, and $\{\alpha_t\}_{t=1}^T$, $\{\beta_t\}_{t=1}^T$ are discrete parameter sequences in DDPM.
However, $\alpha_t$ in $\dd \alpha_t / \dd t$ refers to the continuous-time parameter determined by the following score SDE~\citep{song2020score}
\begin{equation}
\label{eq:sde}
    \dd \rvx(\tau) = - \frac{1}{2}\beta(\tau) \rvx(\tau) \dd \tau + \sqrt{\beta(\tau)} \dd \rvw,~\tau \in [0, 1],
\end{equation}
where $\rvw$ is the standard Wiener process, $\beta(\tau)$ is the continuous version of $\{\beta_t\}_{t=1}^T$ with a continuous time variable $\tau \in [0,1]$ for indexing, and the continuous-time $\alpha_t = \exp{(-\frac{1}{2}\int_{0}^t \beta(s) \dd s)}$.
To avoid ambiguity, let $\alpha(\tau),~\tau\in[0,1]$ denote the continuous version of $\{\alpha_t\}_{t=1}^T$. Thus,
\begin{equation}
\label{eq:alpha-derivative}
    \left. \frac{\dd \alpha(\tau)}{\dd \tau}\right\vert_{\tau=0} = \left. -\frac{1}{2} \beta(\tau) \exp{(-\frac{1}{2}\int_{0}^\tau \beta(s) \dd s)} \right\vert_{\tau=0} = -\frac{1}{2} \beta(0).
\end{equation}
Once the continuous function $\beta(\tau)$ is determined for a specific noise schedule, we can obtain $\left. \frac{\dd \alpha(\tau)}{\dd \tau}\right\vert_{\tau=0}$ immediately by~\cref{eq:alpha-derivative}.

To obtain $\beta(\tau)$, we first give the expression of $\{\beta_t\}_{t=1}^T$ in linear and quadratic schedules~\citep{ho2020denoising}
\begin{align}
    \text{Linear: }&  \beta_t = \frac{\bar{\beta}_{\rm min}}{T} + \left(\frac{\bar{\beta}_{\rm max}}{T} - \frac{\bar{\beta}_{\rm min}}{T}\right) \cdot \frac{t -1}{T - 1}, \label{eq:beta-linear}\\
    \text{Quadratic: }& \beta_t = \left(\sqrt{\frac{\bar{\beta}_{\rm min}}{T}} + \left(\sqrt{\frac{\bar{\beta}_{\rm max}}{T}} - \sqrt{\frac{\bar{\beta}_{\rm min}}{T}}\right) \cdot \frac{t -1}{T - 1}\right)^2,\label{eq:beta-quadratic}
\end{align}
where $\bar{\beta}_{\rm min}$ and $\bar{\beta}_{\rm max}$ are user-defined hyperparameters. Then, we define an auxiliary sequence $\{\bar{\beta}_t=T\beta_t\}_{t=1}^T$. In the limit of $T\rightarrow \infty$, $\{\bar{\beta}_t\}_{t=1}^T$ becomes the function $\beta(\tau)$ indexed by $\tau \in [0, 1]$
\begin{align}
    \text{Linear: }&  \beta(\tau) = \bar{\beta}_{\rm min} + \left(\bar{\beta}_{\rm max} - \bar{\beta}_{\rm min}\right) \cdot \tau, \label{eq:continuous-beta-linear}\\
    \text{Quadratic: }& \beta(\tau) = \left(\sqrt{\bar{\beta}_{\rm min}} + \left(\sqrt{\bar{\beta}_{\rm max}} - \sqrt{\bar{\beta}_{\rm min}} \right) \cdot \tau\right)^2,\label{eq:continuous-beta-quadratic}
\end{align}
Thus, $\beta(0) = \bar{\beta}_{\rm min}$ for both linear and quadratic schedules, which leads to $\left. \frac{\dd \alpha(\tau)}{\dd \tau}\right\vert_{\tau=0} = -\frac{1}{2}\bar{\beta}_{\rm min}$. 
As a common setting, $\bar{\beta}_{\rm min}$ is a positive real number, thus $\left. \frac{\dd \alpha(\tau)}{\dd \tau}\right\vert_{\tau=0} < 0$.

\subsubsection{$\dd \alpha_t / \dd t$ for the cosine schedule at zero point}\label{ssubsec:alpha-derivatives-c}

The cosine schedule is designed to prevent abrupt changes in noise level near $t=0$ and $t=T$~\citep{nichol2021improved}.
Different from linear and quadratic schedules that define $\{\alpha_t\}_{t=1}^T$ by a pre-designed sequence $\{\beta_t\}_{t=1}^T$, the cosine schedule directly defines $\{\alpha_t\}_{t=1}^T$ as
\begin{equation}
\label{eq:alpha-cosine}
    \alpha_t = \frac{f(t)}{f(0)},\quad f(t)=\cos{\left(\frac{t/T + s}{1 + s} \cdot \frac{\pi}{2} \right)},\quad t=1,2,\cdots,T,
\end{equation}
where $s$ is a small positive offset.
The continuous version of $\{\alpha_t\}_{t=1}^T$ can be obtained in the limit of $T\rightarrow \infty$ as
\begin{equation}
\label{eq:alpha-cosine-continuous}
\alpha(\tau) = \cos{\left(\frac{\tau + s}{1 + s} \cdot \frac{\pi}{2} \right)} / \cos{\left(\frac{s}{1 + s} \cdot \frac{\pi}{2} \right)},\quad \tau\in [0,1].
\end{equation}

With \cref{eq:alpha-cosine-continuous}, we can easily get $\left. \frac{\dd \alpha(\tau)}{\dd \tau}\right\vert_{\tau=0}$
\begin{equation}
    \left. \frac{\dd \alpha(\tau)}{\dd \tau}\right\vert_{\tau=0} = -\frac{\pi}{2(1+s)} \tan{\left(\frac{s}{1 + s} \cdot \frac{\pi}{2} \right)},
\end{equation}
which leads to $\left. \frac{\dd \alpha(\tau)}{\dd \tau}\right\vert_{\tau=0} < 0$ since $s > 0$.

\begin{figure*}[t]
\begin{center}
% \begin{overpic}[width=0.99\linewidth]{figures/v-prediction.pdf}
\begin{overpic}[width=0.99\linewidth]{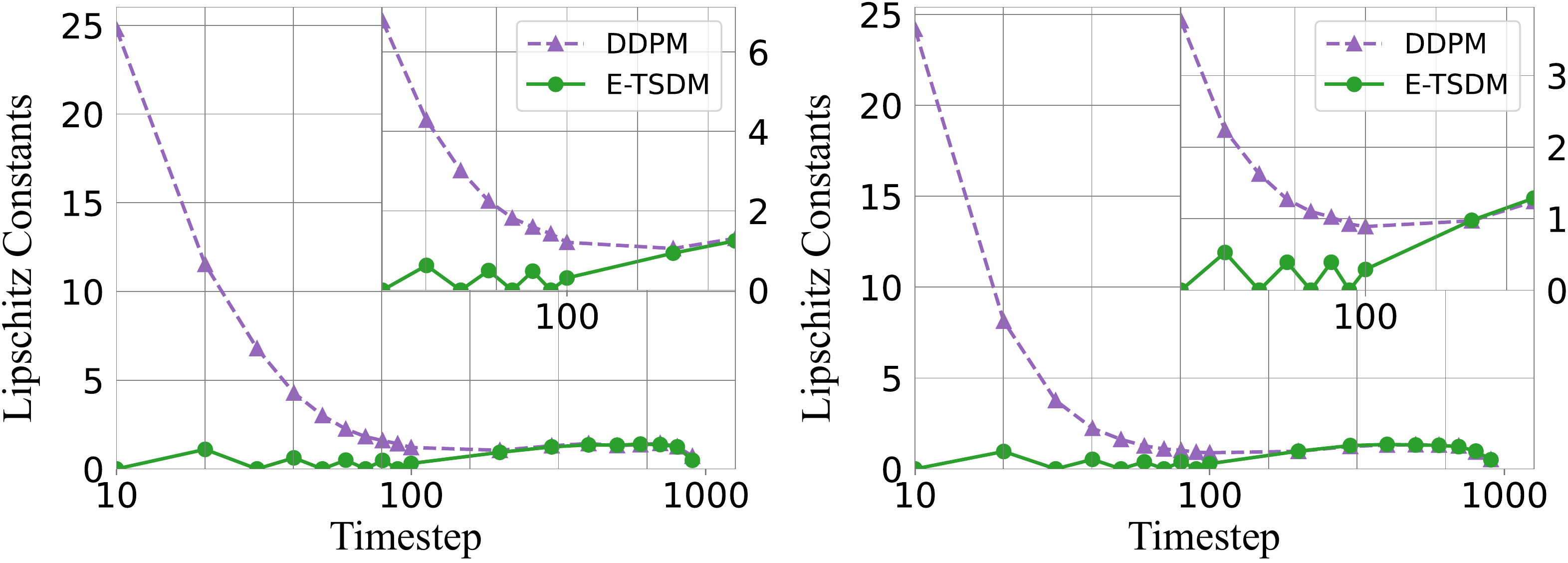}
  \put(15,-3){(a) LSUN-Cat $256\times256$}
  \put(67,-3){(b) FFHQ $256\times256$}
\end{overpic}
\end{center}
\vspace{4pt}
\caption{\textbf{Quantitative comparison} of the Lipschitz constants between \methodabbr and DDPM~\citep{ho2020denoising} using \textbf{v-prediction}~\citep{salimans2022progressive} on Lsun-Cat $256\times256$~\citep{karras2019style} and FFHQ $256\times256$ dataset~\citep{karras2019style}.
Results show that \methodabbr can efficiently reduce the Lipschitz constants in \textbf{v-prediction} scenarios.}
\label{fig:app-v-prediction}
% \vspace{-5pt}
\end{figure*}

\subsection{Lipschitz singularies for v-prediction diffusion models}\label{subsec:v-prediction}

In \cref{sec:analyze} of the main paper, we prove that noise-prediction diffusion models suffer from Lipschitz singularities issue.
In this section, we show that the Lipschitz singularities issue is also an important problem for v-prediction diffusion models from both theoretical and empirical perspectives.

Theoretically, the optimal solution of v-prediction models is 
\begin{equation}
\begin{aligned}
v(\rvx, t) &= \underset{v_\theta}{\operatorname{argmin}} \mathbb{E} [ \| v_\theta(\mathbf{x_t}, t) - (\alpha_t \beps - \sigma_t \mathbf{x}_0) \|^2_2 | \mathbf{x}_t = \rvx] \\\\
&= \mathbb{E}[ \alpha_t \beps - \sigma_t \mathbf{x}_0 | \mathbf{x}_t = \rvx] \\\\
&= \mathbb{E}\left[ \left. \alpha_t \beps - \sigma_t \frac{\mathbf{x}_t - \sigma_t \beps}{\alpha_t} \right| \mathbf{x}_t = \rvx\right] \\\\
&= -\frac{\sigma_t}{\alpha_t} x + (\alpha_t + \frac{\sigma^2_t}{\alpha_t}) \mathbb{E}[\beps | \mathbf{x}_t = \rvx] \\\\
&= -\frac{\sigma_t}{\alpha_t} x - \frac{\alpha^2_t + \sigma^2_t}{\alpha_t} \sigma_t \nabla_x \log q_t(\rvx) \\\\
&= -\frac{\sigma_t}{\alpha_t} (\rvx + \nabla_\rvx \log q_t(\rvx)),
\end{aligned}
\end{equation}
where $\rvx + \nabla_\rvx \log q_t(\rvx)$ is smooth under the assumption of \cref{prop:xt-diff}, and $\frac{\dd}{\dd t}\left(\frac{\sigma_t}{\alpha_t}\right) \rightarrow \frac{\dd \sigma_t}{\dd t}$ as $t \rightarrow 0$.
Thus, with the same derivation of \cref{prop:xt-diff}, we can conclude that $\lim \sup_{t\rightarrow 0^+} \left\|\frac{\partial v(x, t)}{\partial t}\right\|\rightarrow \infty$.
The detailed derivation goes as follows: 

Firstly, we can obtain the partial derivative of the v-prediction model over $t$ as 
\begin{equation}
    \frac{\partial v(\rvx, t)}{\partial t} = -\frac{\dd }{\dd t}(\frac{\sigma_t}{\alpha_t}) (\rvx + \nabla_\rvx \log q_t(\rvx)) - \frac{\sigma_t}{\alpha_t} \frac{\dd }{\dd t} (\rvx + \nabla_\rvx \log q_t(\rvx)).
\end{equation}
Note that $\frac{d}{dt}\left(\frac{\sigma_t}{\alpha_t}\right) = \frac{1}{\alpha_t^2} \left( \alpha_t \frac{\dd \sigma_t}{\dd t} - \sigma_t \frac{\dd \alpha_t}{\dd t}\right) \rightarrow \frac{\dd \sigma_t}{\dd t} = -\frac{\alpha_t}{\sqrt{1 - \alpha_t^2}}\frac{\dd \alpha_t}{\dd t}$ as $t \rightarrow 0$ under common settings that $\sigma_0=0$, $\alpha_0=1$, and $\left.\frac{\dd \alpha_t}{\dd t}\right|_{t = 0}$ is finite, thus if $\left.\frac{\dd \alpha_t}{\dd t}\right|_{t = 0} \neq 0$, and $\rvx + \nabla_\rvx \log q_t(\rvx) \neq \mathbf{0}$, then one of the following two must stand
\begin{equation}
    \lim \sup_{t \rightarrow 0^+} \left\Vert \frac{\partial v(\rvx, t)}{\partial t} \right\Vert \rightarrow \infty;\quad \lim \sup_{t \rightarrow 0^+}  \left\Vert \frac{\sigma_t}{\alpha_t} \frac{\dd }{\dd t} (\rvx + \nabla_\rvx \log q_t(\rvx)) \right\Vert \rightarrow \infty.
\end{equation}
Under the assumption that $q_t(\rvx)$ is a smooth process, we can conclude that $\lim \sup_{t \rightarrow 0^+} \left\Vert \frac{\partial v(\rvx, t)}{\partial t} \right\Vert \rightarrow \infty$.

Since most of the diffusion models are noise-prediction and v-prediction models, the Lipschitz singularities issue is an important problem for the community of diffusion models.

Empirically, we can also observe the phenomenon of Lipschitz singularities for v-prediction diffusion models, where the experimental results of Lipschitz constants on FFHQ $256\times256$ dataset~\citep{karras2019style} and Lsun-Cat $256\times256$~\citep{karras2019style} are shown in \cref{fig:app-v-prediction}, from which we can tell \methodabbr can effectively mitigate Lipschitz singularities in v-prediction scenario.
% %
Besides, we also provide corresponding quantitative evaluations evaluated by FID-10k in \cref{tab:v-prediction}, showing that \methodabbr can also improve the synthesis performance in the v-prediction scenario.

\begin{table*}[t]
\hspace{1pt}
\begin{minipage}[c]{0.42\textwidth}
\centering
\caption{\textbf{Quantitative comparison} between \methodabbr and DDPM~\citep{ho2020denoising} using \textbf{v-prediction} on Lsun-Cat $256\times256$~\citep{karras2019style} and FFHQ $256\times256$ dataset~\citep{karras2019style} evaluated with FID-10k $\downarrow$.
Experimental results indicate that \methodabbr can achieve better synthesis performance.}
\label{tab:v-prediction}
\vspace{3pt}
\centering
\setlength{\tabcolsep}{4pt}
  \begin{tabular}{l cc}
    \toprule
    & Baseline  & \methodabbr \\
    \midrule
    FFHQ   & 10.85  & $\bf9.00$ \\
    Lsun-Cat  & 18.40  & $\bf13.86$ \\
    \bottomrule
  \end{tabular}
\end{minipage}
\hfill
\begin{minipage}[c]{0.49\textwidth}
\centering
\caption{\textbf{Quantitative comparison} among \methodabbr, DDPM~\citep{ho2020denoising}, and DDPM using \textbf{regularization} techniques (DDPM-r) on FFHQ $256\times256$~\citep{karras2019style} and CelebAHQ $256\times256$~\citep{karras2017progressive} evaluated with FID-10k $\downarrow$.
Experimental results show that \textbf{DDPM-r can slightly improve the FID but performs worse than \methodabbr}.
}
\label{tab:regularization}
\vspace{3pt}
\setlength{\tabcolsep}{4pt}
% \small
\centering
  \begin{tabular}{lccc}
    \toprule
    Method  &Baseline  &\methodabbr  &DDPM-r  \\
    \midrule
    FFHQ  &9.50  &$\bf6.62$  &9.18  \\
    CelebAHQ  &8.05  &$\bf6.99$  &7.97  \\
    \bottomrule
  \end{tabular}
% \vspace{-5pt}
\end{minipage}
\hspace{1pt}
\vspace{-7pt}
\end{table*}

% \subsection{The introduced error of \methodabbr can be bounded by an infinitesimal}\label{subsec:bound-proof}
\subsection{Proof of \Cref{prop:bound}}\label{subsec:bound-proof}

Here we will first give the derivation of the upper-bound on $\Vert \beps^*(\rvx, f_{\sT}(t)) - \beps(\rvx, t)\Vert$ when $t \in [0, \tilde{t})$, where $\beps^*(\rvx, f_{\sT}(t))$ denotes the optimal $\beps_\theta(\rvx, f_{\sT}(t))$, and $\beps(\rvx, t)=-\sigma_t \nabla_\rvx \log q_t(\rvx)$. Then, we will discuss the convergence rate of the error bound.

For any $t \in [0, \tilde{t})$, there exists an $i \in \{1, 2, \cdots, n\}$ such that $t \in [t_{i-1}, t_i)$.
For simplicity, we use $h(\rvx, t)$ to denote the score function $\nabla_\rvx \log q_t(\rvx)$, and use $\mathbb{E}_\tau[\cdot]$ to denote the expectation over $\tau \sim \gU(t_{i-1}, t_i)$.
Thus, we can obtain
\begin{equation}
\begin{aligned}
\label{eq:error-bound-derivation}
\left\Vert \beps^*(\rvx, f(t)) - \beps(\rvx, t)\right\Vert &= \left\Vert \mathbb{E}_\tau[\beps(\rvx, \tau)] - \beps(\rvx, t)\right\Vert \\
&=\left\Vert \mathbb{E}_\tau [\sigma_\tau  h(\rvx, \tau)] - \sigma_t h(\rvx, t)\right\Vert \\
&=\left\Vert \mathbb{E}_\tau [\sigma_\tau  h(\rvx, \tau) - \sigma_\tau h(\rvx, t) + \sigma_\tau h(\rvx, t) - \sigma_t h(\rvx, t)]\right\Vert \\
&\leq \left\Vert \mathbb{E}_\tau [\sigma_\tau \left(h(\rvx, \tau) - h(\rvx, t)\right)] \right\Vert + \left\Vert \mathbb{E}_\tau [(\sigma_\tau - \sigma_t) h(\rvx, t)] \right\Vert\\
&\leq \mathbb{E}_\tau [\sigma_\tau \Vert h(\rvx, \tau) - h(\rvx, t)\Vert ] + \mathbb{E}_\tau [|\sigma_\tau - \sigma_t|]\Vert h(\rvx, t) \Vert \\
&\leq \sigma_{t_i} \mathbb{E}_\tau [\Vert h(\rvx, \tau) - h(\rvx, t)\Vert ] + (\sigma_{t_i} - \sigma_{t_{i-1}})\Vert h(\rvx, t) \Vert \\
&\leq \sigma_{t_i} K_i(\rvx) (t_i - t_{i-1}) + B_i(\rvx) (\sigma_{t_i} - \sigma_{t_{i-1}}) \\
&\leq \sigma_{\tilde{t}} K(\rvx) \Delta t + B(\rvx) \Delta \sigma_{\max},
\end{aligned}
\end{equation}

where $K_i(\rvx)=\sup_{t,\tau\in[t_{i-1}, t_i),t\neq \tau} \frac{\Vert h(\rvx,t) - h(\rvx,\tau)\Vert}{|t - \tau|}$, $B_i(\rvx)=\sup_{t\in[t_{i-1}, t_i)} \Vert h(\rvx, t)\Vert$, $K(\rvx)=\sup_{t,\tau\in[0, \tilde{t}),t\neq \tau} \frac{\Vert h(\rvx,t) - h(\rvx,\tau)\Vert}{|t - \tau|}$, $B(\rvx)=\sup_{t\in[0, \tilde{t})} \Vert h(\rvx, t)\Vert$, and $\Delta \sigma_{\max} = \max_{1\leq i\leq n}|\sigma_{t_i} - \sigma_{t_{i-1}}|$. The first equality holds because
\begin{equation}
\begin{aligned}
\beps(\rvx,t)&=\argmin_{\beps_\theta} \mathbb{E}[\Vert \beps_\theta(\mathbf{\rvx}_\tau, \tau) - \beps \Vert_2^2 | \tau=t, \mathbf{\rvx}_\tau=\rvx] \\
&=\mathbb{E}[\beps | \tau=t, \mathbf{\rvx}_\tau=\rvx], 
\end{aligned}
\end{equation}
and our optimal $\beps^*(\rvx, f(t))$ can be expressed as 
\begin{equation}
\begin{aligned}
\beps^*(\rvx, f(t)) &= \beps^*(\rvx, t_{i-1}) \\
&=\argmin_{\beps_\theta} \mathbb{E}_{\tau \sim \gU(t_{i-1}, t_i), \beps} [\Vert \beps_\theta(\mathbf{\rvx}_\tau, t_{i-1}) - \beps \Vert_2^2 | \mathbf{\rvx}_\tau=\rvx] \\
&= \mathbb{E}_{\tau \sim \gU(t_{i-1}, t_i), \beps} [\beps | \mathbf{\rvx}_\tau=\rvx] \\
&= \mathbb{E}_{\tau \sim \gU(t_{i-1}, t_i)} \mathbb{E}_{\beps} [\beps | \tau, \mathbf{\rvx}_\tau=\rvx] \\
&=\mathbb{E}_{\tau \sim \gU(t_{i-1}, t_i)} [\beps(\rvx,\tau)].
\end{aligned}
\end{equation}

As for the rate of convergence, it is obvious from \cref{eq:error-bound-derivation} that we only need to determine the convergence rate of $\Delta \sigma_{\max}$. 
Under common settings, $\sigma_t$ is monotonically decreasing and concave for $t \in [0, T]$, thus
\begin{equation}
    \Delta \sigma_{\max} = \max_{1\leq i\leq n}|\sigma_{t_i} - \sigma_{t_{i-1}}|
    = \sigma_{t_1} - \sigma_{t_0}
    = \sigma_{\Delta t},
\end{equation}
where the last equality holds because $\sigma_{t_0} = \sigma_{0} = 0$, and $t_1 = \tilde{t} / n = \Delta t$ as we uniformly divides $[0, \tilde{t})$ into $n$ sub-intervals. Then, we can verify the convergence rate of $\Delta \sigma_{\max}$ as
\begin{equation}
\begin{aligned}
    \lim_{\Delta t \rightarrow 0} \frac{\Delta \sigma_{\max}}{\sqrt{\Delta t}} &= \lim_{\Delta t \rightarrow 0} \sqrt{\frac{\sigma_{\Delta t}^2}{\Delta t}}\\
    &= \left. \sqrt{\frac{\dd \sigma_t^2}{\dd t} }\right|_{t = 0} \\
    &= \left. \sqrt{\frac{\dd (1 - \alpha_t^2)}{\dd t} }\right|_{t = 0} \\
    &= \left. \sqrt{-2 \alpha_t  \frac{\dd \alpha_t}{\dd t}}\right|_{t = 0} \\
    &= \left. \sqrt{-2 \frac{\dd \alpha_t}{\dd t}}\right|_{t = 0}, 
\end{aligned}
\end{equation}
where $\left.\frac{\dd \alpha_t}{\dd t}\right|_{t = 0}$ is finite and $\left.\frac{\dd \alpha_t}{\dd t}\right|_{t = 0} \leq 0$. 
Thus, we can conclude that $\Delta \sigma_{\max}$ is at least $\frac{1}{2}$-order convergence with respect to $\Delta t$, and the error bound $\sigma_{\tilde{t}} K(\rvx) \Delta t + B(\rvx) \Delta \sigma_{\max}$ is also at least $\frac{1}{2}$-order convergence.
This is a relatively fast convergence speed in optimization, and demonstrates that the introduced errors of \methodabbr are controllable.

\begin{table*}[t]
\hspace{1pt}
\begin{minipage}[c]{0.45\textwidth}
\centering
\caption{\textbf{Quantitative comparison} among \methodabbr, DDPM~\citep{ho2020denoising}, and \textbf{modification of noise schedules (Modified-NS)} on 
FFHQ $256\times256$ dataset~\citep{karras2019style} evaluated with FID-10k $\downarrow$.
Specifically, we implement Modified-NS on linear, quadratic, and cosine schedules.
Experimental results indicate that the performance of \textbf{Modified-NS is unstable while \methodabbr achieves better synthesis performance}.}
\label{tab:moified-noise-schedule}
\vspace{1pt}
\centering
\setlength{\tabcolsep}{4pt}
  \begin{tabular}{lccc}
    \toprule
    & Linear  & Quadratic  & Cosine  \\
    \midrule
    Baseline  & 9.50  & 13.79  & 27.17  \\
    \methodabbr  & $\bf6.62$  & $\bf9.69$  & $\bf26.08$  \\
    Modified-NS  & 8.67  & 17.48  & 26.84 \\
    % & Baseline  & \methodabbr \\
    % \midrule
    % FFHQ   & 10.85  & $\bf9.00$ \\
    % Lsun-Cat  & 18.40  & $\bf13.86$ \\
    \bottomrule
  \end{tabular}
\end{minipage}
\hfill
\begin{minipage}[c]{0.5\textwidth}
\centering
\caption{\textbf{Quantitative comparison} of \textbf{remap} method between uniformly sampling $t$ and uniformly sampling $\lambda$, during training and inference, on FFHQ $256\times256$~\citep{karras2019style} evaluated with FID-10k $\downarrow$.
Specifically, $\mathcal{U}_t$ is $\mathcal{U}[0,1]$, and $\mathcal{U}_\lambda$ is $\mathcal{U}[0, K]$ for $1/t$ but $\mathcal{U}[-K, K]$ for Inverse-Sigmoid, where $K$ is a large number to avoid infinity.
Results show that \textbf{remap is not helpful}.
% Experimental results show that DDPM-r can slight\methodabbr performs better than DDPM-r.
}
\label{tab:remap}
\vspace{5pt}
\setlength{\tabcolsep}{4pt}
% \small
\centering
  \begin{tabular}{l|l|cc}
    \toprule
    Training  &Inference  &\multicolumn{2}{c}{Remap Function} \\
    \cmidrule{3-4}
    Strategy  &Strategy  &$1/t$  &Inverse-Sigmoid  \\
    \midrule
    $t\sim \mathcal{U}_t$  &$t\sim \mathcal{U}_t$  &9.43  &9.33  \\
    $t\sim \mathcal{U}_t$  &$\lambda \sim \mathcal{U}_\lambda$  &83.71  &468.90  \\
    $\lambda \sim \mathcal{U}_\lambda$  &$t\sim \mathcal{U}_t$  &83.44  &468.19  \\
    $\lambda\sim \mathcal{U}_\lambda $  &$\lambda\sim \mathcal{U}_\lambda $  &171.06  &351.89 \\
    \bottomrule
  \end{tabular}
% \vspace{-5pt}
\end{minipage}
\hspace{1pt}
\vspace{-7pt}
\end{table*}

\section{Additional results}\label{sec:app-results}

\subsection{Lipschitz constants}\label{sec:app-results:lipschitz-constants}

In our main paper, we demonstrate the effectiveness of \methodabbr in reducing the Lipschitz constants near $t=0$ by comparing its Lipschitz constants with that of DDPM baseline~\citep{ho2020denoising} on the FFHQ $256\times256$ dataset~\citep{karras2019style}.
As a supplement, we provide additional comparisons of Lipschitz constants on other datasets, including AFHQ-Cat~\citep{choi2020stargan} (see \cref{fig:supp-lipschitz-constants}a), AFHQ-Wild~\citep{choi2020stargan} (see \cref{fig:supp-lipschitz-constants}b), Lsun-Cat $256\times256$~\citep{karras2019style} (see \cref{fig:supp-lipschitz-constants}c), Lsun-Church $256\times256$~\citep{karras2019style} (see \cref{fig:supp-lipschitz-constants}d), and CelebAHQ $256\times256$~\citep{karras2017progressive} (see \cref{fig:supp-lipschitz-constants}e).
These experimental results demonstrate that \methodabbr is highly effective in mitigating Lipschitz singularities in diffusion models across various datasets.

\begin{figure*}[t]
\hspace{2pt}
\begin{minipage}[c]{0.46\textwidth}
\centering
% \vspace{-2pt}
% \includegraphics[width=1\textwidth]{figures/lipschitz_constants_ffhq_cosine_shift.pdf}
\begin{overpic}[width=0.99\linewidth]{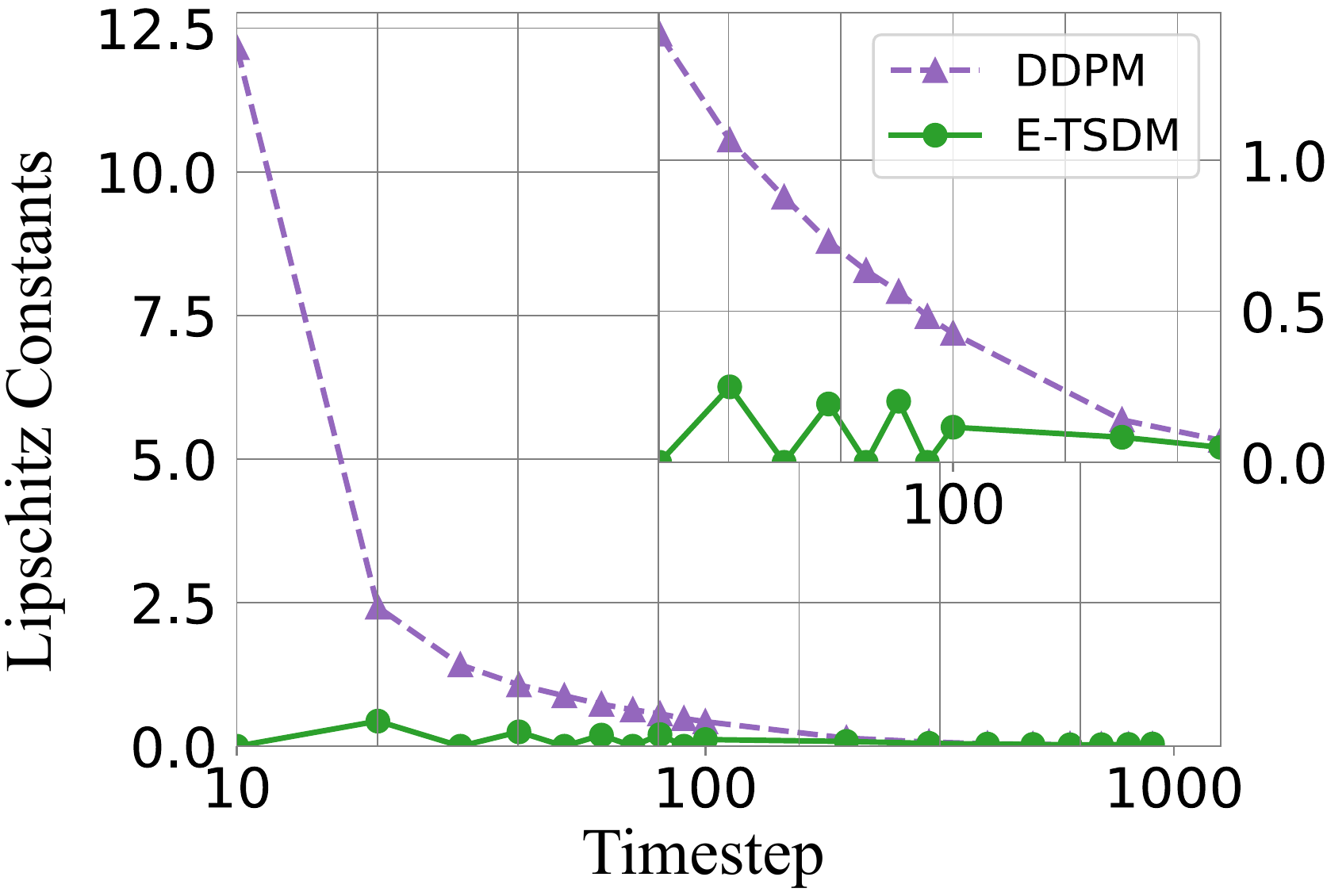}
\end{overpic}
% \vspace{1pt}
\caption{\textbf{Quantitative comparison} of Lipschitz constants between \methodabbr and DDPM baseline~\citep{ho2020denoising} on FFHQ $256\times256$~\citep{karras2019style} using the \textbf{cosine shift schedule}.}
\vspace{-10pt}
\label{fig:lipschitz-cosine-shift}
\end{minipage}
\hfill
\begin{minipage}[c]{0.46\textwidth}
\centering
% \vspace{-2pt}
% \includegraphics[width=1\textwidth]{figures/lipschitz_constants_ffhq_difference.pdf}
\begin{overpic}[width=0.99\linewidth]{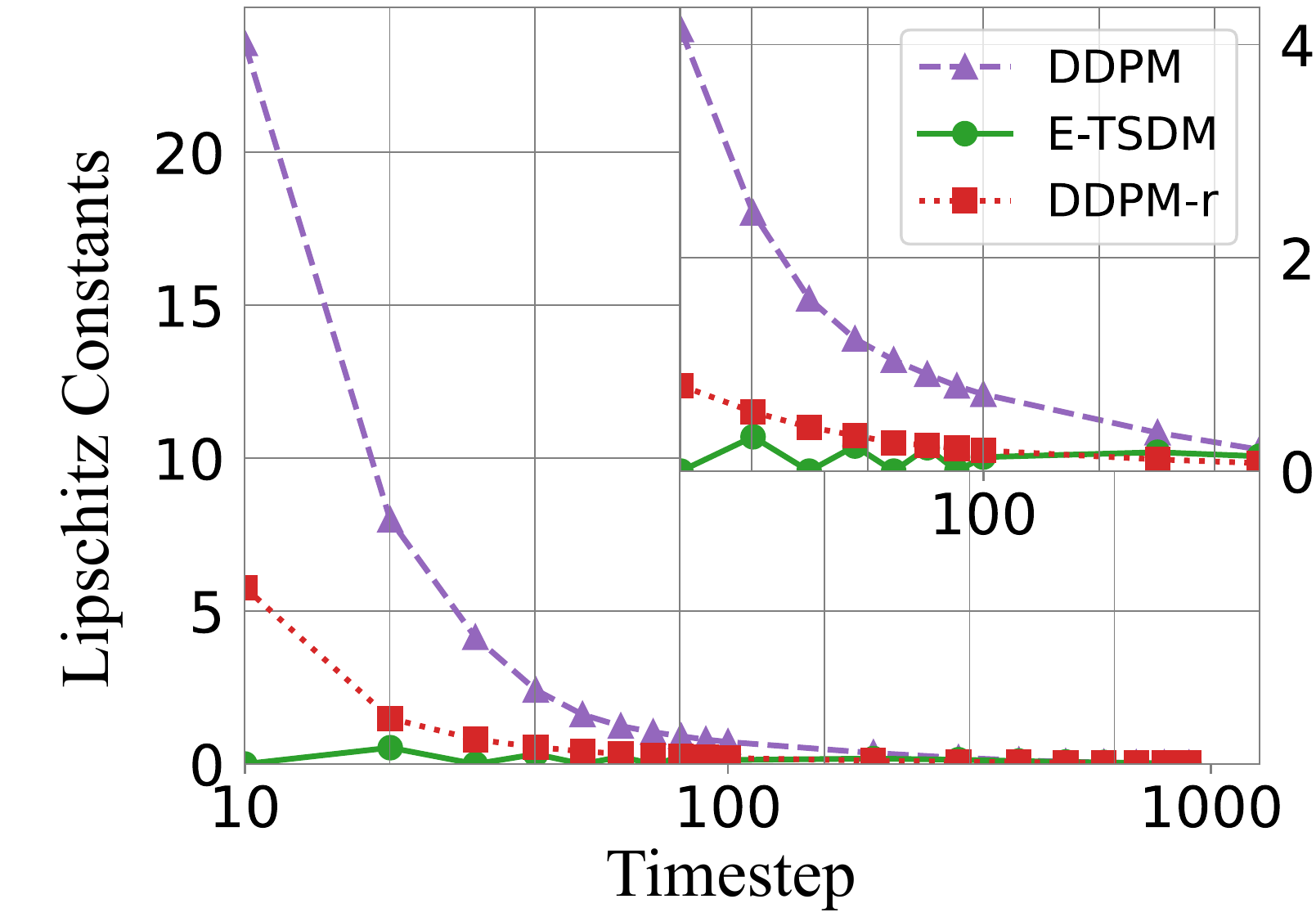}
  % \put(42,-8){Timestep}
  % \put(15,-5){0}
  % \put(8.5,2){0}
  % \put(8.5,62){24}
  % \put(57,-5){100}
  % \put(90,-5){1000}
  % \put(2,11){\rotatebox{90}{Lipschitz Constants}}
\end{overpic}
% \vspace{1pt}
\caption{\textbf{Quantitative comparison} of Lipschitz constants among \methodabbr, DDPM~\citep{ho2020denoising}, and DDPM~\citep{ho2020denoising} using \textbf{regularization} techniques (DDPM-r) on FFHQ $256\times256$~\citep{karras2019style}.}
\vspace{-10pt}
\label{fig:lipschitz-difference}
\end{minipage}
\hspace{5pt}
\end{figure*}

Furthermore, we provide a comparison of Lipschitz constants between \methodabbr and the DDPM baseline~\citep{ho2020denoising} when using the quadratic schedule and the cosine-shift schedule~\citep{hoogeboom2023simple}.
As shown in \cref{fig:supp-lipschitz-constants}f, we observe that large Lipschitz constants still exist in diffusion models when using the quadratic schedule, and \methodabbr effectively alleviates this problem.
Similar improvement can also be observed when using the cosine-shift schedule as illustrated in \cref{fig:lipschitz-cosine-shift}, highlighting the superiority of our approach over the DDPM baseline.

\subsection{Quantitative analysis of $\tilde{t}$ and $n$}\label{sec:app-results:quantitative and ablation:ablation}

In our main paper, we investigated the impact of two important settings for \methodabbr, the length of the interval to share conditions $\tilde{t}$, and the number of sub-intervals $n$ in this interval.
As a supplement, we provide additional results on various datasets to further investigate the optimal settings for these parameters.

As seen in \cref{fig:supp-ab-t} and \cref{fig:supp-ab-n}, we observe divergence in the best choices of $n$ and $\tilde{t}$ across different datasets.
However, we find that the default settings where $\tilde{t}=100$ and $n=5$ consistently yield good performance across a range of datasets. 
Based on these findings, we recommend the default settings as an ideal choice for implementing \methodabbr without the need for a thorough search.
However, if performance is the main concern, researchers may conduct a grid search to explore the optimal values of $\tilde{t}$ and $n$ for specific datasets.

% \todo{}

\subsection{Alternative methods}\label{sec:alternative-methods}

In this section, we discuss three different alternative methods that possibly alleviate Lipschitz singularities. including regularization, modification of noise schedules, and remap.
Although seem feasible, they have different problems, resulting in worse performance than \methodabbr.

\subsubsection{Regularization}\label{sec:alternative-methods:regularization}

As mentioned in the main paper, one alternative method is to impose restrictions on the Lipschitz constants through regularization techniques.
In this section, we apply regularization on the baseline and estimate the gradient of $\beps_\theta(\rvx, t)$ by calculating the difference $K(t, t^\prime)$.
We represent this method as DDPM-r in this paper.
As shown in \cref{fig:lipschitz-difference}, although DDPM-r can also reduce the Lipschitz constants, its capacity to do so is substantially inferior to that of \methodabbr.
Additionally, DDPM-r necessitates twice the calculation compared to \methodabbr.
Regarding synthesis performance, as shown in \cref{tab:regularization}, DDPM-r performs slightly better than baseline, but much worse than \methodabbr, indicating that \methodabbr is a better choice than regularization.

\begin{figure*}[t]
\hspace{2pt}
\begin{minipage}[c]{0.46\textwidth}
\centering
% \vspace{-2pt}
% \includegraphics[width=1\textwidth]{figures/lipschitz_constants_ffhq_cosine_shift.pdf}
\begin{overpic}[width=0.99\linewidth]{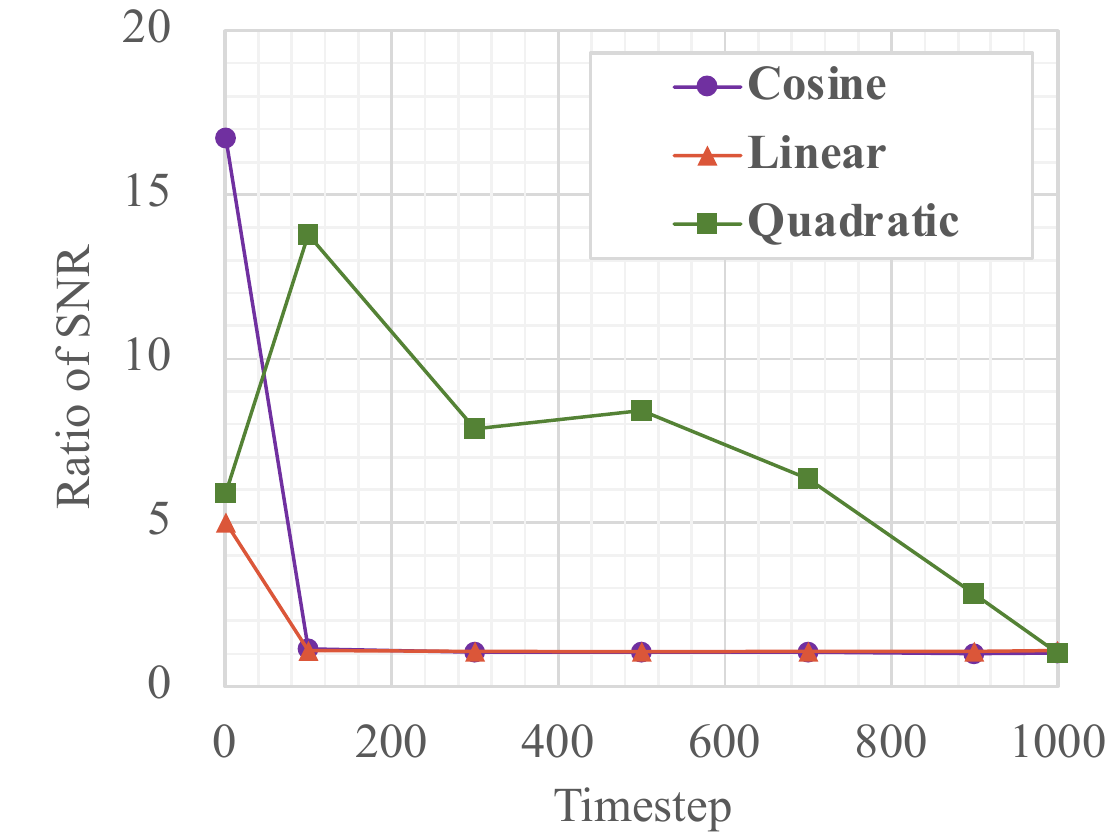}
  % \put(45,-8){Timestep}
  % \put(15,-5){0}
  % \put(5,1.5){1}
  % \put(4.5,64){20}
  % \put(54,-5){500}
  % \put(90,-5){1000}
  % \put(1,20){\rotatebox{90}{Ratio of SNR}}
\end{overpic}
% \vspace{-6pt}
\caption{\textbf{Quantitative evaluation} of the \textbf{ratio of SNR} of Modified-NS to the SNR of the corresponding original noise schedule.
Results show that \textbf{Modified-NS} significantly increases the SNR near zero point, and thus reduces the amounts of added noise near zero point.
Specifically, for the quadratic schedule, Modified-NS seriously increases the SNR almost during the whole process.}
\vspace{-10pt}
\label{fig:snr_modified_ns}
\end{minipage}
\hfill
\begin{minipage}[c]{0.46\textwidth}
\centering
% \vspace{4pt}
% \includegraphics[width=1\textwidth]{figures/lipschitz_constants_ffhq_difference.pdf}
\begin{overpic}[width=0.99\linewidth]{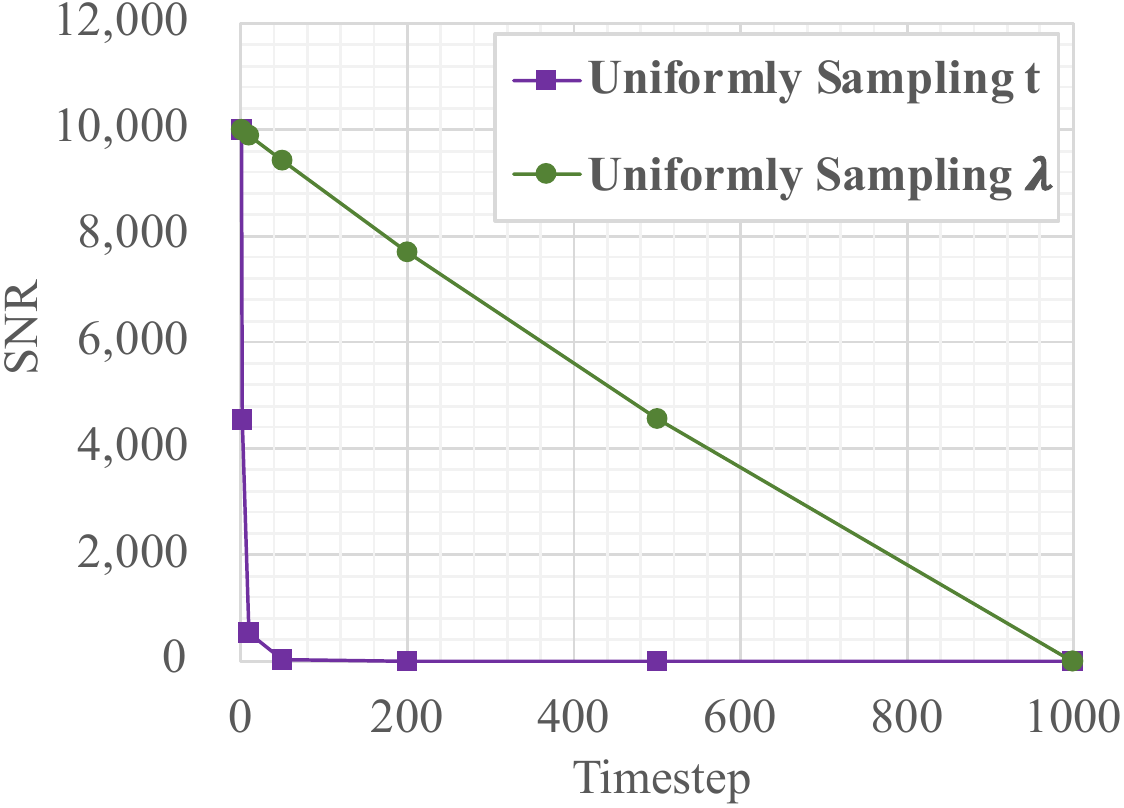}
\end{overpic}
% \vspace{0pt}
\caption{\textbf{Quantitative comparison} of SNR for remap method between uniformly sampling $t$ and uniformly sampling the remapped conditional input $\lambda$.
Results show that when using \textbf{remap} method, uniformly sampling $\lambda$ significantly increases the SNR across all of the timesteps, and thus forces the network to focus too much on the beginning stage of the diffusion process.}
\vspace{-10pt}
\label{fig:snr_remap}
\end{minipage}
\hspace{5pt}
\end{figure*}

\subsubsection{Modifying noise schedules}\label{sec:alternative-methods:noise-schedules}

As proved in \cref{sec:app-derivation}, the mainstream noise schedules satisfy $\left. \frac{\dd\alpha_t}{\dd t} \right\vert_{t=0} \neq 0$, leading to Lipschitz singularities as proved in \cref{prop:xt-diff}.
However, it is possible to modify those schedules to force them to have $\left. \frac{\dd\alpha_t}{\dd t} \right\vert_{t=0} = 0$, and thus alleviate Lipschitz singularities.
We represent this method as Modified-NS in this paper.
However, as said in~\citet{nichol2021improved}, $\left. \frac{\dd\alpha_t}{\dd t} \right\vert_{t=0} = 0$ means tiny amounts of noise at the beginning of the diffusion process, making it hard for the network to predict accurately enough.

To explore the performance, we conduct experiments of Modified-NS on FFHQ $256\times256$~\citep{karras2019style} for all of the three discussed noise schedules in \cref{subsec:alpha-derivatives}.
Specifically, for linear and quadratic schedules, since $\left. \frac{\dd \alpha(\tau)}{\dd \tau}\right\vert_{\tau=0} = -\frac{1}{2} \beta(0)$ (as detailed in \cref{eq:alpha-derivative}), we implement Modified-NS by setting $\beta(0)=0$.
Note that for the quadratic schedule, such a modification will significantly magnify the Signal to Noise Ratio (SNR), $\frac{\alpha_t^2}{\sigma_t^2}$, across the whole diffusion process, so we slightly increase $\beta_T$ to make its SNR at $t=T$ similar to that of the original quadratic schedule.
Meanwhile, $\beta_1, \dots, \beta_{T-1}$ are also correspondingly increased due to $\beta_t = (\sqrt{\beta_0} + (\sqrt{\beta_T}-\sqrt{\beta_0})\frac{t}{T-1})^2$.
As for the cosine schedule, we set the offset $s$ in \cref{eq:alpha-cosine} to zero.
Experimental results are shown in \cref{tab:moified-noise-schedule}, from which we find that the performance of Modified-NS is unstable.
More specifically, Modified-NS improves performance for linear and cosine schedules but significantly drags down the performance for the quadratic schedule.
We further provide the comparison of SNR between Modified-NS and their corresponding original noise schedules in \cref{fig:snr_modified_ns} by calculating the ratio of Modified-NS's SNR to the original noise schedule's SNR.
From this figure we can tell that for linear and cosine schedule, Modified-NS significantly increase the SNR near zero point while maintaining the SNR of other timesteps similar.
In other words, on the one hand, Modified-NS seriously reduces the amount of noise added near zero point, which can be detrimental to the accurate prediction.
On the other hand, Modified-NS alleviates the Lipschitz singularities, which is beneficial to the synthesis performance.
As a result, for linear and cosine schedules, Modified-NS performs better than baseline but worse than \methodabbr.
However, for the quadratic schedule, although we force the SNR of Modified-NS at $t=T$ similar to the SNR of the original schedule, the SNR at other timesteps is significantly increased, leading to a worse performance of Modified-NS compared to that of baseline.

\subsubsection{Remap}\label{sec:alternative-methods:remap}

Except for regularization and Modified-NS, remap is another possible method to fix the Lipschitz singularities issue.
Recall that the inputs of network $\beps_\theta(\rvx, t)$ is noisy image $\rvx$ and timestep condition $t$.
Remap is trying to design a remap function $\lambda=f(t)$ on $t$ as the conditional input of network instead of $t$, namely, $\beps_\theta(\rvx, f(t))$.
The core idea of remap is to reduce $\frac{\partial \beps_\theta(\rvx, t)}{\partial t}$ by significantly stretching the interval with large Lipschitz constants.
Note that although $f_{\sT}$ of \methodabbr can also be seen as a kind of remap function, there are big differences between \methodabbr and remap.
Specifically, \methodabbr tries to set the numerator to zero while remap aims to significantly increase the denominator.
Besides, $f_{\sT}$ has no inverse while $f(t)$ of remap is usually a reversible function.
We provide two simple choices of $f(t)$ in this section as examples, which are $f(t) = 1/t$ and $f^{-1}(\lambda) = \text{sigmoid}(\lambda)$.

Remap can efficiently reduce the Lipschitz constants regarding the conditional inputs of the network, $\frac{\partial \beps_\theta(\rvx, t)}{\partial \lambda}$.
However, since we uniformly sample $t$ both in training and inference, what should be focused on is the Lipschitz constants regarding $t$, $\frac{\partial \beps_\theta(\rvx, t)}{\partial t}$, which can not be influenced by remap.
In other words, although remap seems to be a feasible method, it is not helpful to mitigate the Lipschitz constants we care about, unless we uniformly sample $\lambda$ in training and inference.
However, uniformly sampling $\lambda$ may force the network to focus on a certain part of the diffusion process.
We use $f(t)=1/t$ as an example to illustrate this point and show the comparison of SNR between uniformly sampling $t$ and uniformly sampling $\lambda$ when using remap in \cref{fig:snr_remap}.
Results show that uniformly sampling $\lambda$ maintains a high SNR across all of the timesteps, leading to excessive attention to the beginning stage of the diffusion process.
As a result, when we uniformly sample $\lambda$ during training or inference, the synthesis performance gets significantly worse as shown in \cref{tab:remap}.
Besides, when we uniformly sample $t$ both in training and inference, remap makes no difference and thus leads to a similar performance to the baseline.

\begin{figure*}[t]
\hspace{2pt}
\begin{minipage}[c]{0.52\textwidth}
\centering
% \vspace{-2pt}
\begin{overpic}[width=0.99\linewidth]{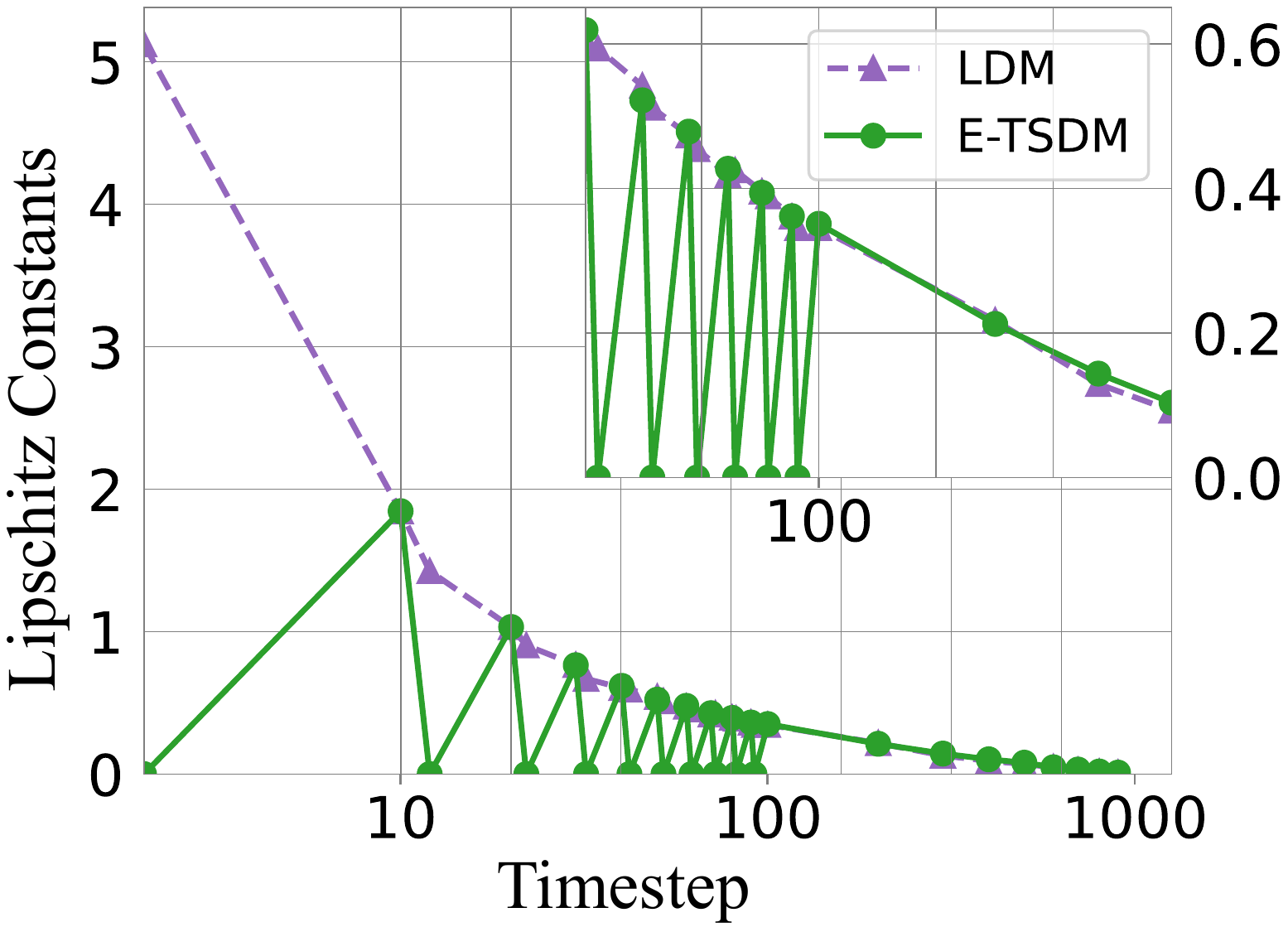}
\end{overpic}
% \vspace{2pt}
\caption{\textbf{Quantitative comparison} of Lipschitz constants between \methodabbr and LDM~\citep{rombach2022high} on FFHQ $256\times256$\citep{karras2019style}.
\methodabbr reduces the overall Lipschitz constants near $t=0$, and mitigates the Lipschitz singularities occurring in LDM~\citep{rombach2022high}.
}
% \vspace{-10pt}
\label{fig:ldm-lipschitz}
\end{minipage}
\hfill
\begin{minipage}[c]{0.40\textwidth}
\centering
% \vspace{-10pt}
\includegraphics[width=1\textwidth]{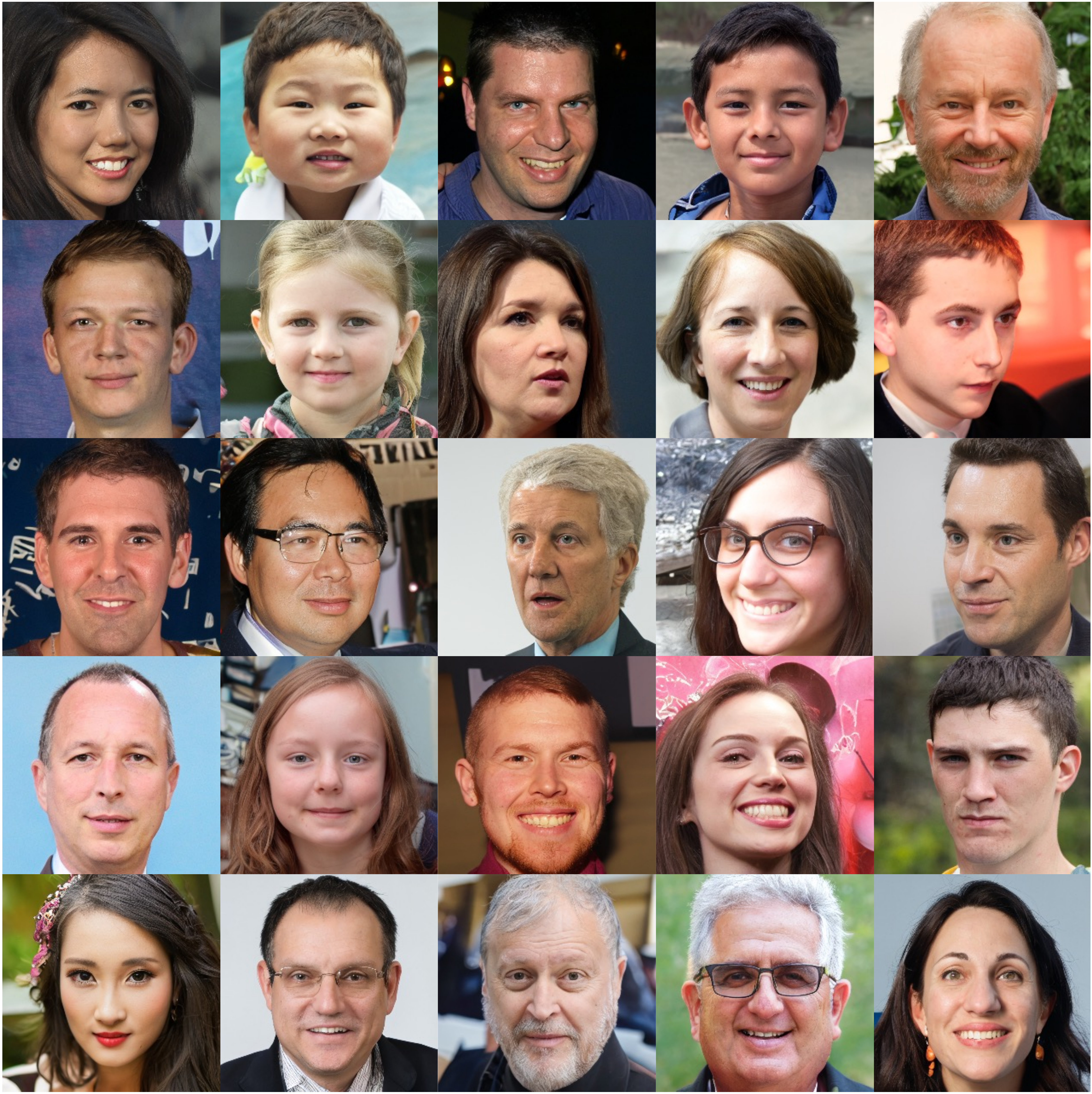}
\caption{\textbf{Qualitative results} produced by \methodabbr implemented on LDM~\citep{rombach2022high} on FFHQ $256\times256$\citep{karras2019style}.}
% \vspace{-5pt}
\label{fig:ldm-sample}
\end{minipage}
\hspace{5pt}
\end{figure*}

\begin{figure*}[t]
\centering
\begin{overpic}[width=1.0\linewidth]{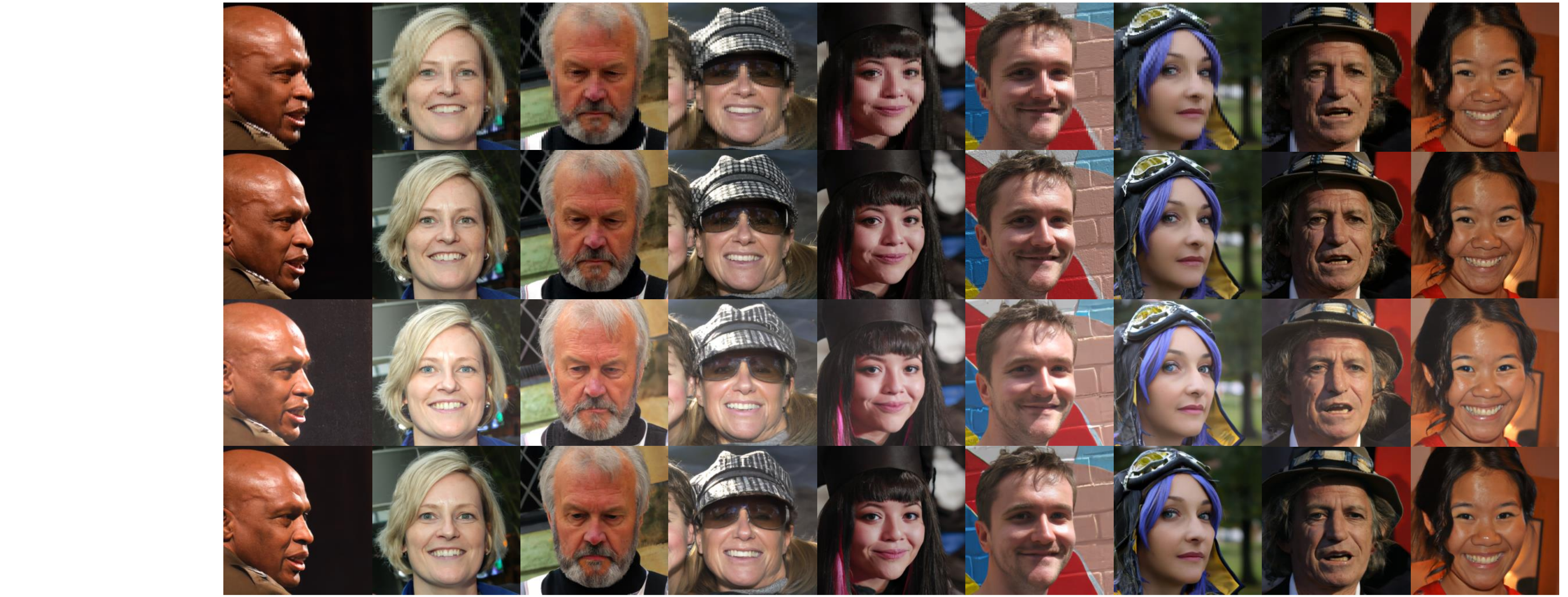}
% \fbox{\rule[-.5cm]{0cm}{4cm} \rule[-.5cm]{13cm}{0cm}}
\put(-2,32){Low resolution}
\put(-0.5,23){Original one}
\put(1,14.5){Baseline}
\put(-1,12){PSNR$\uparrow$: 24.64}
\put(2,4.5){Ours}
\put(-1,2){PSNR$\uparrow$: 25.61}
\end{overpic}
\vspace{-15pt}
\caption{
    \textbf{Qualitative and quantitative results} by applying \methodabbr to super-resolution task (\textit{i.e.}, from $64\times64$ to $256\times256$), using PSNR as the evaluation metric. 
    Results show that \methodabbr mitigates the color bias occurring in baseline and improves the PSNR from 24.64 to 25.61, which suggests that \textbf{our approach well supports conditional generation}.
}
\label{fig:super-resolution}
\vspace{-10pt}
\end{figure*}

\subsection{More diffusion models}

Latent diffusion models (LDM)~\citep{rombach2022high} is one of the most renowned variants of diffusion models.
In this section, we will investigate the Lipschitz singularities in LDM~\citep{rombach2022high}, and apply \methodabbr to address this problem.
LDM~\citep{rombach2022high} shares a resemblance with DDPM~\citep{rombach2022high} but has an additional auto-encoder to encode images into the latent space.
As LDM typically employs the quadratic schedule, it is also susceptible to Lipschitz singularities, as confirmed in \cref{fig:ldm-lipschitz}.

As seen in \cref{fig:ldm-lipschitz}, by utilizing \methodabbr, the Lipschitz constants within each timestep-shared sub-interval are reduced to zero, while the timesteps located near the boundaries of the sub-intervals exhibit a Lipschitz constant comparable to that of baseline, leading to a decrease in overall Lipschitz constants in the target interval $t \in[0, \tilde{t})$, where $\tilde{t}$ is set as the default, namely $\tilde{t}=100$.
Consequently, \methodabbr achieves an improvement in FID-50k from 4.98 to 4.61 with the adoption of \methodabbr, when $n=20$.
We provide some samples generated by the \methodabbr implemented on LDM in \cref{fig:ldm-sample}.

Besides, we also implement our \methodabbr to Elucidated diffusion models (EDM)~\citep{karras2022elucidating}, which proposed several changes to both the sampling and training processes and achieves impressive performance.
Specifically, we reproduce EDM and repeat it three times on CIFAR10 $32\times32$~\citep{krizhevsky2009learning} to get a FID-50k of 1.904 $\pm$ 0.015, which is slightly worse than the official released one.
Then we apply \methodabbr to EDM and repeat it three times to get a FID-50k of 1.797 $\pm$ 0.016, indicating that \methodabbr is also helpful to EDM.

\subsection{Generated samples}\label{app:more-samples}

\begin{figure*}[t]
\begin{center}
% \fbox{\rule[-.5cm]{0cm}{12cm} \rule[-.5cm]{16cm}{0cm}}
\includegraphics[width=1\textwidth]{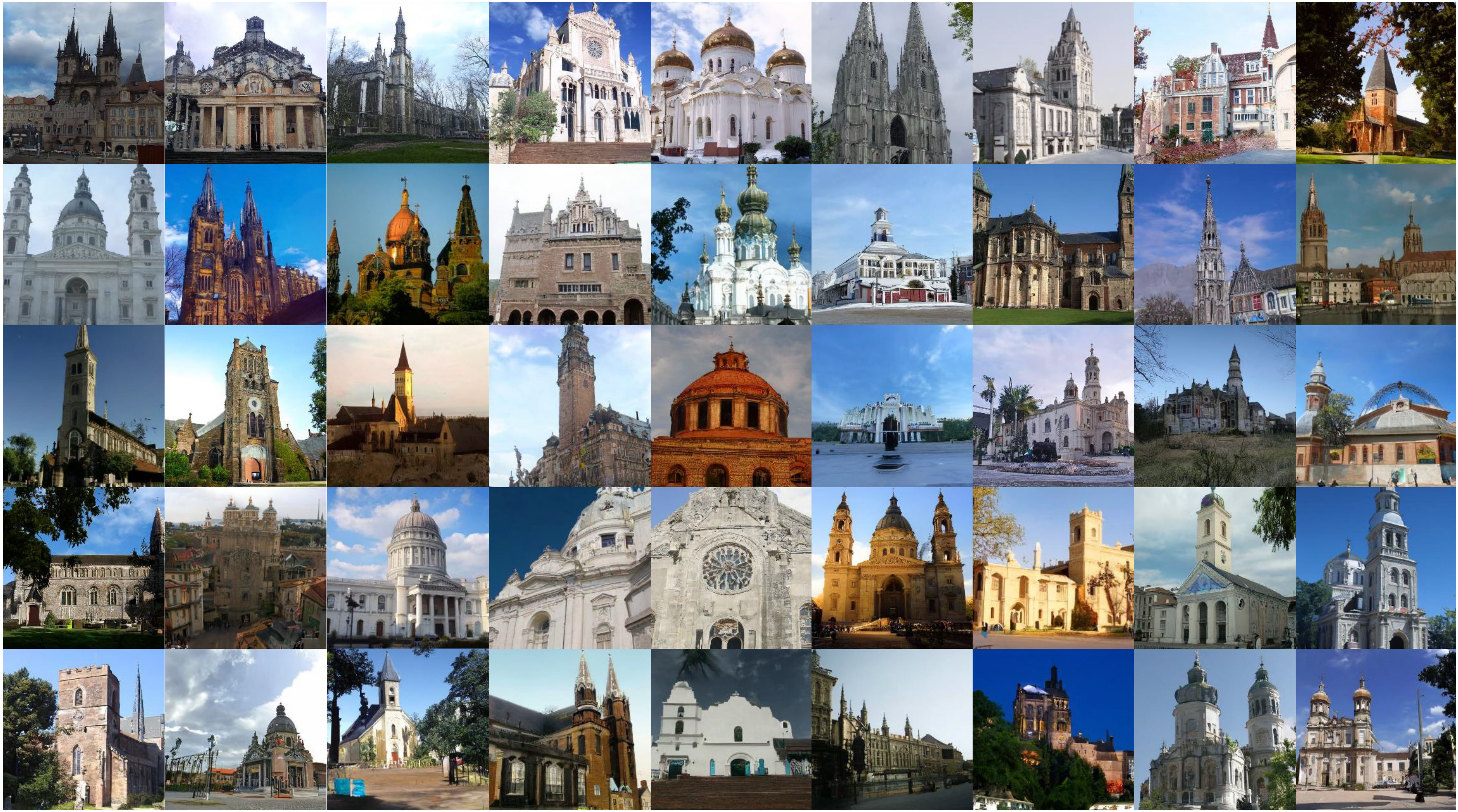}
\end{center}
% \vspace{-15pt}
\caption{\textbf{Qualitative results} produced by \methodabbr on Lsun-Church $256\times256$~\citep{yu2015lsun}.
}
\label{fig:app-lsun-church}
% \vspace{-5pt}
\end{figure*}

\begin{figure*}[t]
\begin{center}
% \fbox{\rule[-.5cm]{0cm}{12cm} \rule[-.5cm]{16cm}{0cm}}
\includegraphics[width=1\textwidth]{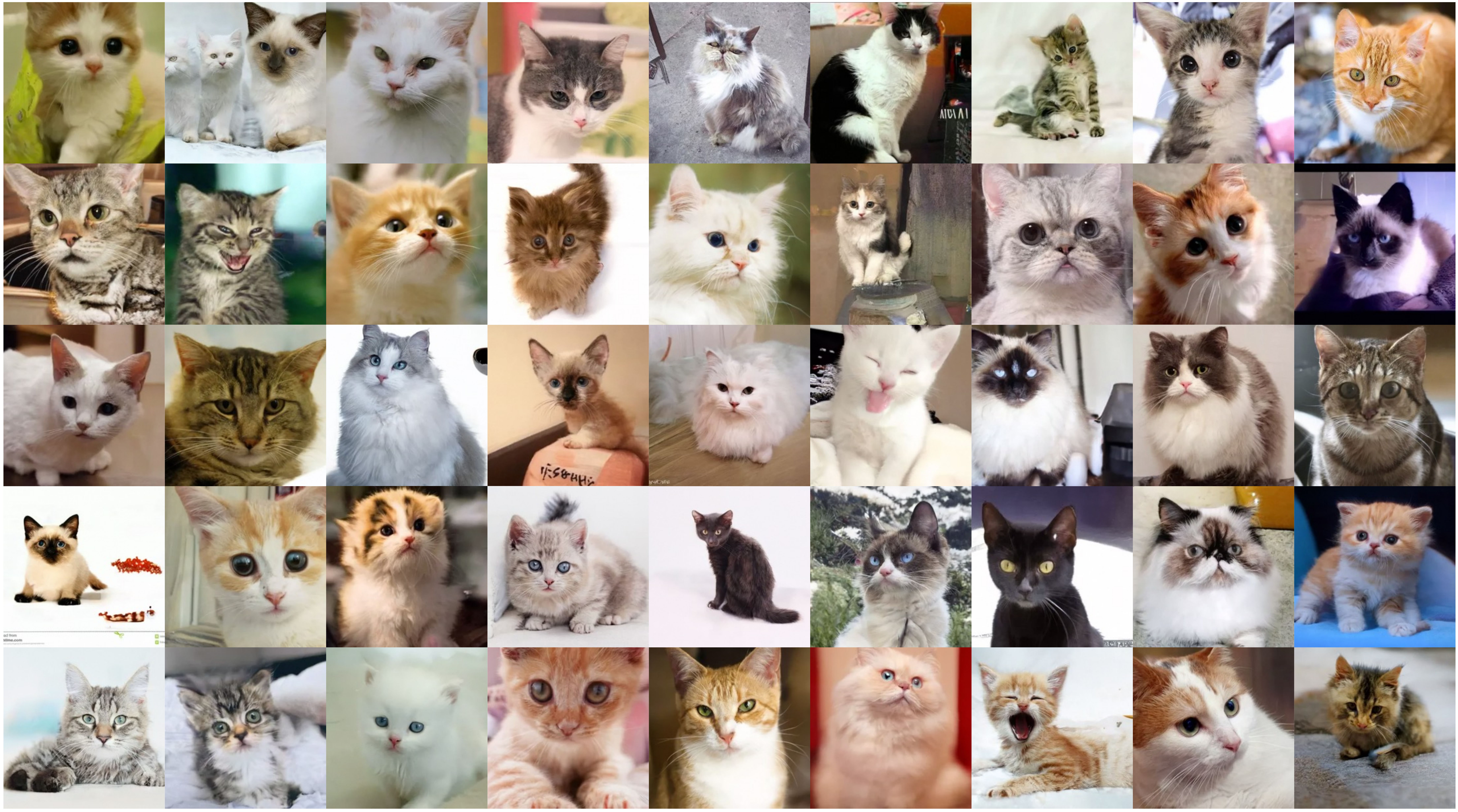}
\end{center}
% \vspace{-15pt}
\caption{\textbf{Qualitative results} produced by \methodabbr on Lsun-Cat $256\times256$~\citep{yu2015lsun}.
}
\label{fig:app-lsun-cat}
% \vspace{-5pt}
\end{figure*}

\begin{figure*}[t]
\begin{center}
% \fbox{\rule[-.5cm]{0cm}{12cm} \rule[-.5cm]{16cm}{0cm}}
\includegraphics[width=1\textwidth]{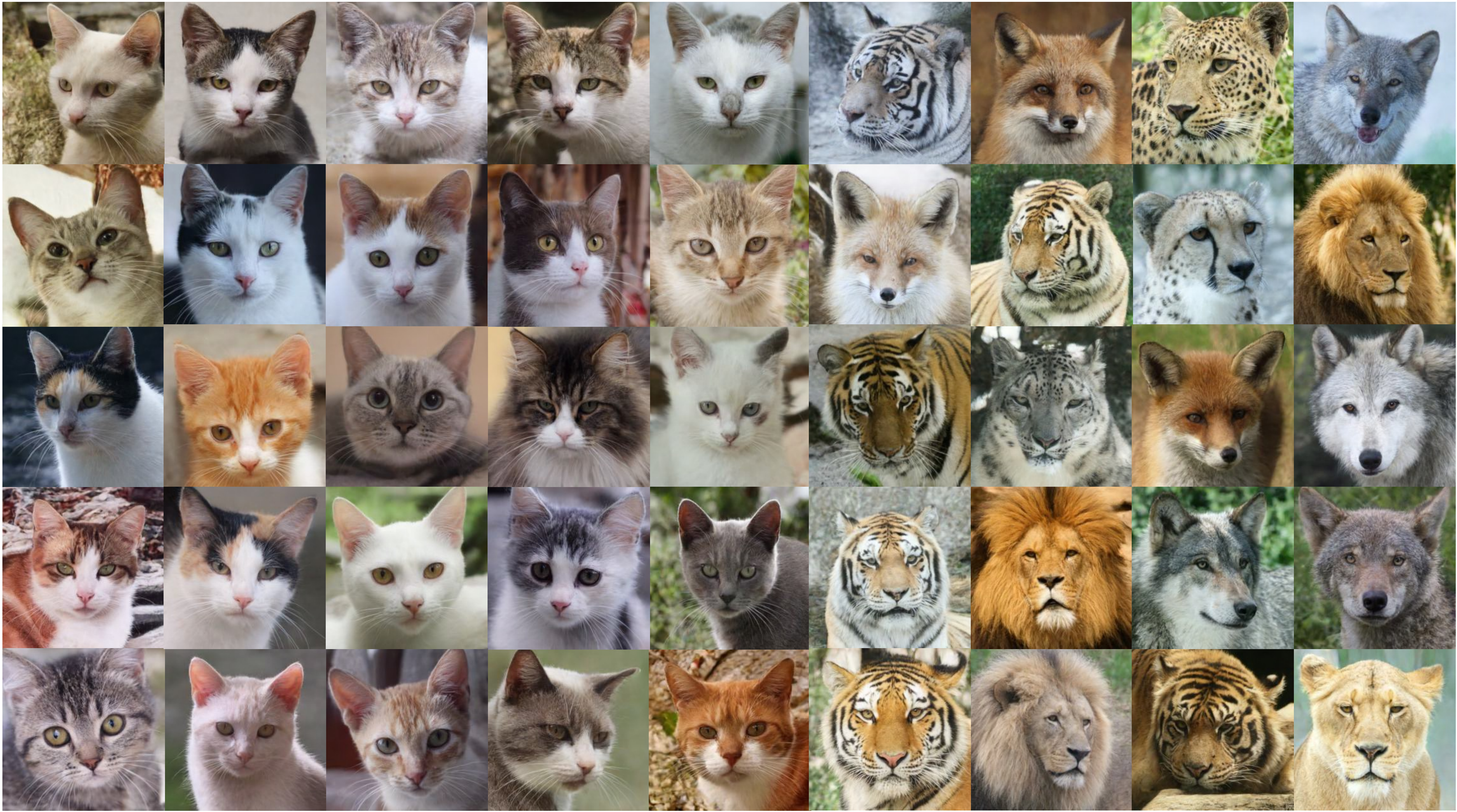}
\end{center}
% \vspace{-15pt}
\caption{\textbf{Qualitative results} produced by \methodabbr on AFHQ-Cat $256\times256$~\citep{choi2020stargan} and AFHQ-Wild $256\times256$~\citep{choi2020stargan}.
}
\label{fig:app-afhqcat}
% \vspace{-5pt}
\end{figure*}

\begin{figure*}[t]
\begin{center}
% \fbox{\rule[-.5cm]{0cm}{12cm} \rule[-.5cm]{16cm}{0cm}}
\includegraphics[width=1\textwidth]{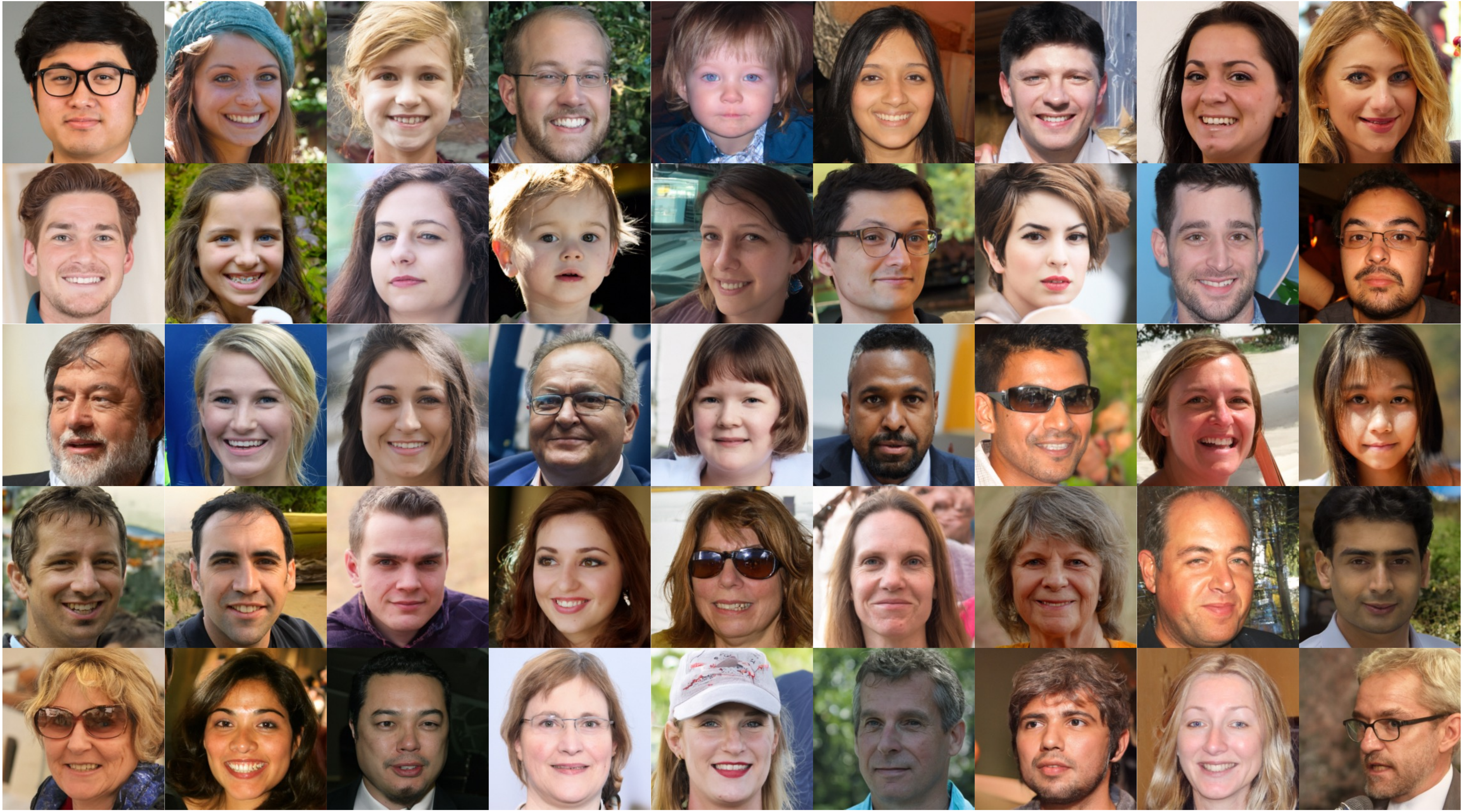}
\end{center}
% \vspace{-15pt}
\caption{\textbf{Qualitative results} produced by \methodabbr on FFHQ $256\times256$\citep{karras2019style}.
}
\label{fig:app-ffhq}
% \vspace{-5pt}
\end{figure*}

\begin{figure*}[t]
\begin{center}
% \fbox{\rule[-.5cm]{0cm}{12cm} \rule[-.5cm]{16cm}{0cm}}
\includegraphics[width=1\textwidth]{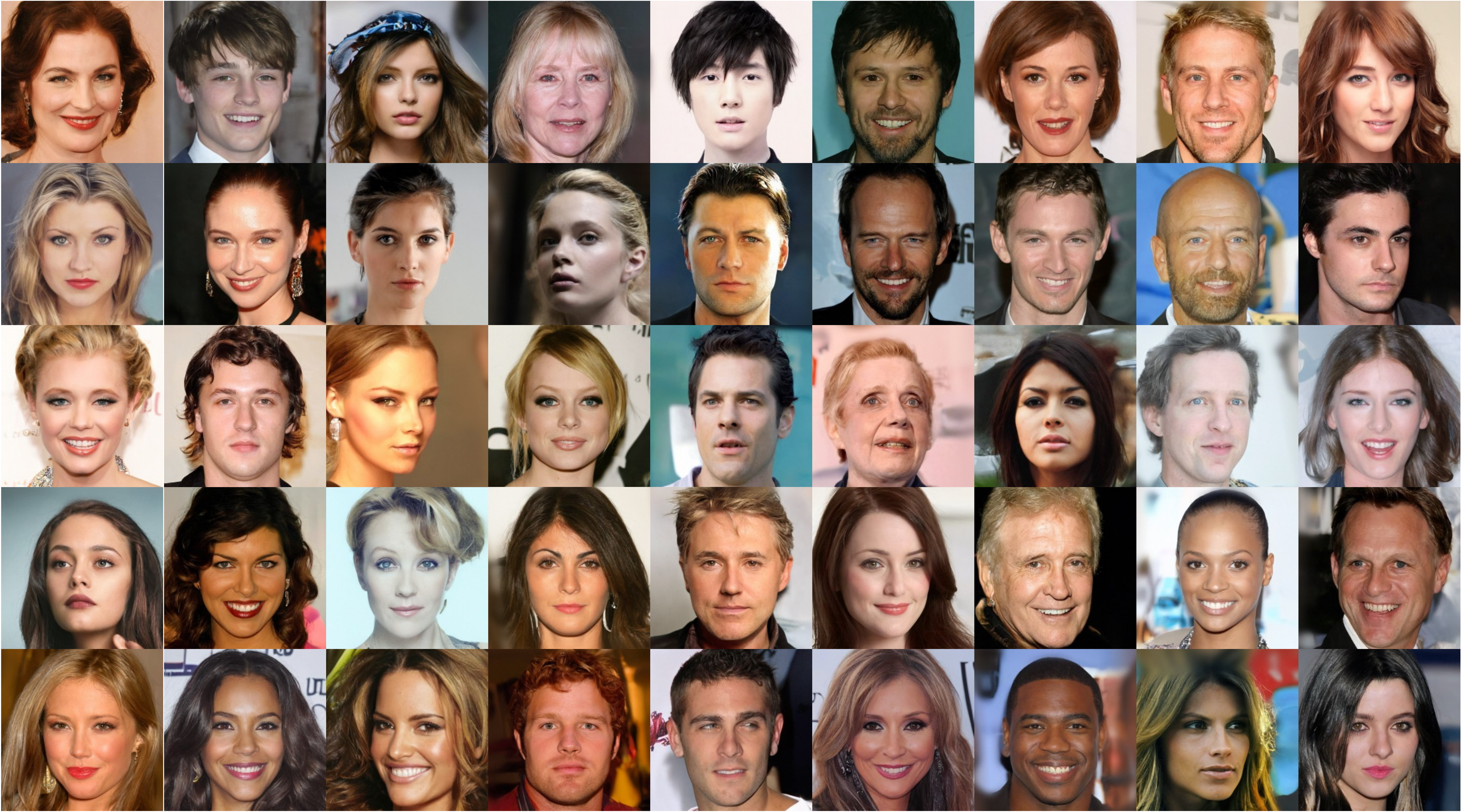}
\end{center}
% \vspace{-15pt}
\caption{\textbf{Qualitative results} produced by \methodabbr on CelebAHQ $256\times256$~\citep{karras2017progressive}.
}
\label{fig:app-celebahq}
% \vspace{-5pt}
\end{figure*}

As a supplement, we provide massive generated samples of \methodabbr trained on Lsun-Church $256\times256$~\citep{karras2019style} (see \cref{fig:app-lsun-church}), Lsun-Cat $256\times256$~\citep{karras2019style} (see \cref{fig:app-lsun-cat}), AFHQ-Cat $256\times256$~\citep{choi2020stargan}, AFHQ-Wild $256\times256$~\citep{choi2020stargan} (see \cref{fig:app-afhqcat}), FFHQ $256\times256$~\citep{karras2019style} (see \cref{fig:app-ffhq}), and CelebAHQ $256\times256$~\citep{karras2017progressive} (see \cref{fig:app-celebahq}).

\end{document}